\documentclass[10pt,twocolumn,letterpaper]{article}

\usepackage{cvpr}
\usepackage{times}
\usepackage{epsfig}
\usepackage{graphicx}
\usepackage{amsmath}
\usepackage{amssymb}
\usepackage[toc,page]{appendix}

\usepackage{algorithm}
\usepackage{algorithmic}
\usepackage{enumitem}
\def\ie{i.e.}
\usepackage[flushleft]{threeparttable} 
\usepackage{multirow}
\usepackage{subfigure}
\usepackage[pagebackref=true,breaklinks=true,letterpaper=true,colorlinks,bookmarks=false]{hyperref}

\def\ie{i.e.}
\def\eg{e.g.}
\def\st{\textrm{s.t.}}
\def\and{\textrm{and}}

\def\b{\textbf{b}}
\def\0{\textbf{0}}
\def\1{\textbf{1}}

\def\b{\boldsymbol{b}}
\def\c{\boldsymbol{c}}
\def\e{\boldsymbol{e}}

\def\q{\boldsymbol{q}}

\def\x{\boldsymbol{x}}

\def\II{\mathcal{I}}
\def\I{\mathbf{I}}

\def\EE{\mathbb{E}}

\def\J{\mathcal{J}}

\newcommand{\RR}{I\!\!R} 

\newcommand{\myparagraph}[1]{\noindent\textbf{#1.}}
\newcommand{\mysubparagraph}[1]{\noindent-- \emph{#1:}}

\newtheorem{theorem}{Theorem}

\newtheorem{proof}{Proof}

\cvprfinalcopy 

\def\cvprPaperID{1850} 

\ifcvprfinal\fi

\begin{document}

\title{Stochastic Sparse Subspace Clustering}

\author{Ying Chen$^1$, Chun-Guang Li$^1$,~and~Chong You$^2$\\
$^1$ SICE, Beijing University of Posts and Telecommunications \\
$^2$ EECS, University of California, Berkeley
}

\maketitle

\thispagestyle{empty}

\begin{abstract}

State-of-the-art subspace clustering methods are based on self-expressive model, which represents each data point as a linear combination of other data points. By enforcing such representation to be sparse, sparse subspace clustering is guaranteed to produce a subspace-preserving data affinity where two points are connected only if they are from the same subspace. On the other hand, however, data points from the same subspace may not be well-connected, leading to the issue of over-segmentation. We introduce dropout to address the issue of over-segmentation, which is based on randomly dropping out data points in self-expressive model. In particular, we show that dropout is equivalent to adding a squared $\ell_2$ norm regularization on the representation coefficients, therefore induces denser solutions. Then, we reformulate the optimization problem as a consensus problem over a set of small-scale subproblems. This leads to a scalable and flexible sparse subspace clustering approach, termed Stochastic Sparse Subspace Clustering, which can effectively handle large scale datasets. Extensive experiments on synthetic data and real world datasets validate the efficiency and effectiveness of our proposal.

\end{abstract}

\section{Introduction}
\label{sec:intro}

In many real world applications, high-dimensional data can be well approximated by a union of low-dimensional subspaces where each subspace corresponds to a class or a category.
The problem of segmenting a set of data points according to the subspaces they belong to, known as \emph{subspace clustering} \cite{Vidal:SPM11-SC, Vidal:Springer16}, has found many important applications such as motion segmentation~\cite{Costeira:IJCV98, Chen:IJCV09}, image clustering~\cite{Lu:ECCV12}, hybrid system identification~\cite{Vidal:ACC04, Bako-Vidal:HSCC08}, matrix completion~\cite{Eriksson:AISTATS12,Li:TSP16}, genes expression clustering~\cite{McWilliams:DMKD14} and so on.

\myparagraph{Prior work}
A traditional method for subspace clustering is $k$-subspaces, which is based on parameterizing a set of basis to the subspaces and finding a segmentation that minimizes the distance of the data points to its corresponding subspaces \cite{Bradley:JGO00, Agarwal:ACM04}.
The $k$-subspaces method requires an accurate estimation of the dimension of the underlying subspaces which is not available in many applications.
In addition, the associated optimization problem is nonconvex, for which a good initialization is important for finding the optimal solution \cite{Lipor:arxiv17,Lane:ICCV19-CoRe}.
Due to the limitations of the $k$-subspaces methods, modern subspace clustering resorts to \emph{spectral clustering} which recovers the segmentation of data from a proper data affinity graph that captures whether two points are from the same subspace or not.
A plethora of early methods for constructing the affinity graph are based on fitting and comparing local subspaces \cite{Yan:ECCV06,Zhang:IJCV12}.
Such methods require dense samples on the subspaces and cannot handle cases where the subspaces are intersecting.

In the past few years, \emph{self-expressive model} 
\cite{Elhamifar:CVPR09,Elhamifar:TPAMI13} has emerged as a powerful tool for computing affinity graph in subspace clustering and has spurred substantial developments and applications.
Given a data matrix $X = [\x_1, \cdots, \x_N] \in \RR^{D\times N}$ whose columns are drawn from a union of subspaces,
self-expressive model states that each data point $\x_j \in \RR^{D}$ can be expressed as a linear combination of other data points, \ie,
\begin{align}
\label{eq:self-expression}
\begin{split}
\x_j =X \c_j + \e_j, ~~~ c_{jj}=0,
\end{split}
\end{align}
where $\c_j \in \RR^N$ is a coefficient vector and $\e_j$ is an error term.
While the linear equation in \eqref{eq:self-expression} may have many feasible solutions, there exists at least one $\c_j$ that is \emph{subspace-preserving}---that is, $c_{ij} \neq 0$ only if points $\x_i$ and $\x_j$ are in the same subspace \cite{Soltanolkotabi:AS12,You:ICML15,Vidal:Springer16}.
Given subspace-preserving representations $[\c_1, \cdots, \c_N]$, the affinity graph is induced 
by an affinity (weight) matrix whose $i, j$-th entry is $|c_{ij}|+|c_{ji}|$.

\myparagraph{Sparse subspace clustering}
Many methods have been proposed for computing subspace-preserving representations by imposing a prior or regularization on the coefficients $\c_j$ \cite{Elhamifar:TPAMI13,Lu:ECCV12,Dyer:JMLR13,Liu:TPAMI13,Soltanolkotabi:AS14,Li:CVPR15,You:CVPR16-SSCOMP,Yang:ECCV16,You:CVPR16-EnSC,Li:TIP17}.
Among them, sparse subspace clustering (SSC) 
\cite{Elhamifar:TPAMI13,You:CVPR16-SSCOMP} that are based on finding the \emph{sparsest} solution to \eqref{eq:self-expression} have become extreme popular due to their theoretical guarantees and empirical success.
Under mild conditions, SSC is guaranteed to recover subspace-preserving solutions even when data points are corrupted with outliers, noises or missing values and when the subspaces are intersecting or affine \cite{Soltanolkotabi:AS12, You:ICML15, Wang:JMLR16, Tsakiris:ICML18, Li:JSTSP18, You:ICCV19}.

While subspace-preserving recovery guarantees that no two points from different subspaces are connected in the affinity graph, 
there is no guarantee that points from the same subspace form a single connected component.
Thus, 
a \emph{connectivity issue} arises that spectral clustering produces an over-segmentation for subspaces with data points 
that are not well-connected. 
In particular, an early work \cite{Nasihatkon:CVPR11} shows that the connectivity issue indeed exists in 
SSC when the dimension of the subspace is greater than $3$.

Several works have attempted to address the connectivity issue in SSC. Motivated by the fact that a low-rank regularization on the matrix of coefficients induces dense solutions, a mixture of $\ell_1$ and nuclear norm is proposed in \cite{Wang:NIPS13-LRR+SSC} to address the connectivity issue.
Unfortunately, solving the optimization problem in \cite{Wang:NIPS13-LRR+SSC} requires doing singular value decomposition in each iteration of the algorithm, which is computationally prohibitive for large scale data. More recently, in \cite{Wang:AISTAT16} a post-processing step that merges potential over-segmented fragments of a subspace into the same cluster is proposed.
While such an approach is conceptually simple and has theoretical guarantees, it only works under the idealized setting where the affinity graph is perfectly subspace-preserving.

\myparagraph{Paper Contributions}
We exploit \emph{dropout} to address the connectivity issue associated with SSC.
\emph{Dropout} is a technique developed for deep learning as an implicit regularization that can effectively alleviate overfitting~\cite{Srivastava:JMLR14,Wan:ICML13,Wager:NIPS13, Baldi:NIPS13, Gal:ICML16, Cavazza:AISTATS18}.
In this paper, dropout refers to the operation of dropping out columns of $X$ uniformly at random when computing the self-expressive representation in \eqref{eq:self-expression}.
Such an operation is equivalent to adding an $\ell_2$ regularization term on the representation coefficient vector $\c_j$, which effectively induces denser solutions. By dropping out columns of the dictionary we solve optimization problems that only involve a (typically very small) part of the original dataset.
This is a particularly attractive property when dealing with ultra-large scale datasets that cannot be loaded into memory.

The contributions of the paper are highlighted as follows.
\vspace{-16pt}
\begin{enumerate}[leftmargin=*]
\item We introduce a \emph{dropout} technique into self-expressive model for subspace clustering, 
and show that it is asymptotically equivalent to a squared $\ell_2$ norm regularizer. 

\vspace{-4pt}
\item We propose a \emph{stochastic sparse subspace clustering} model that is based on dropping out columns of the data matrix. The model has flexible scalability and implicit ability to improve the affinity graph connectivity.

\item We reformulate the stochastic sparse subspace clustering model as a consensus optimization problem and develop an efficient consensus algorithm for solving it.

\vspace{-4pt}
\item We conduct extensive experiments on both synthetic data and real world benchmark data, and demonstrate the state-of-the-art performance of our proposal.

\end{enumerate}

\section{Related Work}

\myparagraph{Self-expressive models in subspace clustering}
Existing subspace clustering methods that are based on self-expressive model 
can be categorized into three groups.
a) For the purpose of inducing subspace-preserving solutions,  existing methods use different regularizations on $\c_j$.
This includes the $\ell_1$ norm \cite{Elhamifar:CVPR09}, the nuclear norm \cite{Liu:ICML10}, the $\ell_2$ norm \cite{Lu:ECCV12}, the traceLasso norm \cite{Lu:ICCV13-TraceLasso},
the $\ell_1$ plus nuclear norms \cite{Wang:NIPS13-LRR+SSC},
the $\ell_1$ plus $\ell_2$ norms in \cite{You:CVPR16-EnSC},
the $\ell_0$ norm in \cite{Yang:ECCV16}
and the weighted $\ell_1$ norm in \cite{Li:CVPR15, Li:TIP17}.
b) To handle different forms of noise that arise in practical applications, existing methods use different regularizations on $\e_j$, \eg, 
the $\ell_1$ and $\ell_2$ norms used in \cite{Elhamifar:CVPR09}, the $\ell_{2,1}$ norm used in \cite{Liu:ICML10},
the mixture of Gaussians in \cite{Li:CVPR15MoG},
and the weighted error entropy proposed in \cite{Li:CVPR19-subspace}.
%
%
c) To perform subspace clustering in an appropriate feature space, self-expressive models are combined with feature learning methods that are based on learning a linear projection \cite{Liu:ICCV11, Patel:ICCV13, Peng:CVPR17} or convolution neural networks \cite{Ji:NIPS17, Zhou:CVPR18, Zhang:CVPR19}.

\myparagraph{Scalable subspace clustering}
In recent years, several attempts to address the scalability of subspace clustering have been proposed.  
For example, in \cite{Peng:CVPR13}, a small subset of data are clustered at first and then the rest of the data are classified based on the learned clusters;
in~\cite{You:CVPR16-SSCOMP, Dyer:JMLR13}, a greedy algorithm
~\cite{Pati:ASILOMAR93} is adopted to solve the sparse self-expression model;
in \cite{Traganitis:TSP18}, a sketching technique is used to speed up SSC;
in \cite{You:Asilomar16}, a divide-and-conquer framework is proposed for extending SSC to large-scale data;
in \cite{Shen:ICML16}, an online dictionary learning based method is proposed to scale up low-rank subspace clustering \cite{Vidal:PRL14, Liu:ICML10};
in \cite{Abdolali:SP19}, SSC is conducted on a hierarchically clustered multiple subsets of the data and then merged via a multi-layer graphs fusion method;
in~\cite{You:ECCV18}, a greedy exemplar selection approach is proposed to extend SSC to handle class-imbalanced data.
While these methods perform subspace clustering on dataset of larger size, there is neither any theoretical guarantee on the quality of the dictionary used in \cite{Abdolali:SP19, Peng:CVPR13, Shen:ICML16} for the purpose of subspace clustering, nor any effort to resolve the connectivity issue of SSC in \cite{Dyer:JMLR13, You:CVPR16-SSCOMP, You:Asilomar16, Traganitis:TSP18, Abdolali:SP19}. As a result, the clustering accuracy in these methods is heavily sacrificed due to using sub-sampled data or erroneous over-segmentation.
%
%
Lastly, almost all the subspace clustering methods mentioned above need to load the entire data into memory. If the size of the data is too large, none of these methods still work.


\section{Dropout in Self-Expressive Model}
\label{sec:dropout-in-columns-of-dictionary}

We formally introduce the dropout operation to the self-expressive model, and show that it is equivalent to adding an $\ell_2$ regularization on the representation vector. 
In the next section, we use such property of dropout to develop a scalable and flexible 
subspace clustering model for addressing the graph connectivity issue associated with SSC.

Consider the problem of minimizing the self-expressive residual as follows:
\begin{equation}
\min \limits_{\c_j} \left\| \x_j - X \c_j \right\|_2^2, \quad \st~~~ c_{jj}=0.
\label{eq:self-expression-noisy-x-j}
\end{equation}

Inspired by the dropout technique used in training neural networks~\cite{Srivastava:JMLR14, Wan:ICML13, Wager:NIPS13, Baldi:NIPS13, Gal:ICML16, Cavazza:AISTATS18}, we propose a dropout operation in the self-expressive model in \eqref{eq:self-expression-noisy-x-j}. 
Similar to dropping ``hidden neurons'' in a neural network, our operation is to discard columns of $X$ uniformly at random.

Specifically, we introduce $0 \le \delta \le 1$ as the dropout rate and let $\{ \xi_{i}\}_{i=1}^N$ be $N$ i.i.d. Bernoulli random variables with probability distribution given by 
\begin{align}
\begin{split}
\xi_{i} =
\begin{cases}
\frac{1}{1-\delta} \quad &\text{with probability} ~~ 1-\delta, \\
0 \quad &\text{with probability}~~\delta.
\end{cases}
\end{split}
\label{eq:xi-definition-Bernoulli}
\end{align}
%
%
Then, dropping the columns of $X$ uniformly at random with probability $\delta$ in \eqref{eq:self-expression-noisy-x-j} is achieved by multiplication of the $N$ i.i.d. Bernoulli random variables $\{ \xi_{i}\}_{i=1}^N$ to the corresponding columns in $X$, \ie,
\begin{equation}
\min \limits_{\c_j} \|  \x_j - \sum_i \xi_i c_{ij} \x_i \|_2^2 \quad \st \quad  c_{jj} =0.
\label{eq:dropout-in-self-expression-noisy-x-j}
\end{equation}

The following theorem gives the asymptotic effect of the dropout in the self-expressive model \eqref{eq:dropout-in-self-expression-noisy-x-j}.
\begin{theorem}
Let $\{ \xi_{i}\}_{i=1}^N$ be $N$ i.i.d. Bernoulli random variables with distribution as defined in \eqref{eq:xi-definition-Bernoulli}. We have that:
\begin{equation}
\begin{split}
&\EE \| \x_j - \sum_i \xi_i c_{ij} \x_i \|_2^2\\
&= \| \x_j - \sum_i c_{ij} \x_i \|_2^2 + \frac{\delta}{1-\delta} \sum_i \|\x_i\|_2^2 c_{ij}^2.
\end{split}
\label{eq:LS-dropout}
\end{equation}
\label{lemma:dropout-LS}
\end{theorem}
\vspace{-15pt}
By Theorem \ref{lemma:dropout-LS}, we can see that the optimization problem
\begin{equation}
\min \limits_{\c_j} \EE \| \x_j - \sum_i \xi_i c_{ij} \x_i \|_2^2   \quad \st \quad  c_{jj} =0,
\label{eq:expectation-dropout-self-expression}
\end{equation}
is equivalent to the optimization problem
\begin{equation}
\begin{split}
\!\!\!\!\min \limits_{\c_j} \!\| \x_j \!- \!\!\!\sum_i c_{ij} \x_i \|_2^2 + \!\frac{\delta}{1-\delta}\!\sum_i \| \x_i \|^2_2 c_{ij}^2~~\st~c_{jj} =0.
\end{split}
\label{eq:expectation-dropout-self-expression-Equivalent}
\end{equation}
In particular, if the columns of $X$ have unit $\ell_2$ norm (\eg, by a data preprocessing step), then  \eqref{eq:expectation-dropout-self-expression-Equivalent} reduces to
\begin{equation}
\min \limits_{\c_j} \| \x_j - \sum_i c_{ij} \x_i \|_2^2 + \lambda \| \c_j \|^2_2  \quad \st~~~c_{jj} =0,
\label{eq:expectation-dropout-self-expression-Equivalent-unit-L2-norm}
\end{equation}
where $\lambda = \frac{\delta}{1-\delta}$.
This is precisely the formulation of the subspace clustering method based on least squares regression \cite{Lu:ECCV12}, and is known to yield dense solutions in general.

In this paper, we aim to develop 
a scalable and flexible subspace clustering method based on the formulation in \eqref{eq:expectation-dropout-self-expression} which,
by means of its equivalency to \eqref{eq:expectation-dropout-self-expression-Equivalent-unit-L2-norm},
has an implicit $\ell_2$ regularization that induces denser solutions.
For practical purpose, we replace the expectation $\EE[\cdot]$ with the \emph{sample mean}, and approach the problem in \eqref{eq:expectation-dropout-self-expression} by solving the following optimization problem
\begin{equation}
\min \limits_{ \c_j } \frac{1}{T} \sum_{t=1}^T \| \x_j - \sum_i \xi_i^{(t)} c_{ij} \x_i \|_2^2   \quad \st~~~c_{jj} =0,
\label{eq:expectation-dropout-self-expression-sample-T}
\end{equation}
where $\xi_i^{(t)}$ is the $t$-th instance of the Bernoulli random variable drawn independently from the distribution 
in \eqref{eq:xi-definition-Bernoulli}.
%
%
%
%

\section{Stochastic Sparse Subspace Clustering: Formulation and A Consensus Algorithm}
\label{sec:dropout-meet-OMP-Stochastic-SSC}

As briefly discussed in the introduction, sparse subspace clustering aims to find a self-expressive representation with the sparest coefficient vector.
That is, it aims to solve the following optimization problem
\begin{equation}
\min \limits_{\c_j} \left\| \x_j - X \c_j \right\|_2^2,~~~\text{s.t.} ~~\|\c_j\|_0 \le s, ~~c_{jj}=0,
\label{eq:self-expression-omp}
\end{equation}
where $\|\cdot\|_0$ is the $\ell_0$ pseudo-norm that counts the number of nonzero entries in the vector and $s$ is a tuning parameter that controls the sparsity of the solution.
It has been shown in \cite{You:CVPR16-SSCOMP} that under mild conditions, the greedy algorithm known as Orthogonal Matching Pursuit (OMP) \cite{Pati:ASILOMAR93} for solving \eqref{eq:self-expression-omp} provably produces a subspace-preserving solution.
On the other hand, it is also established in \cite{You:CVPR16-SSCOMP} that the number of nonzero entries in a subspace-preserving solution produced by OMP cannot exceed the dimension of the subspace that $\x_j$ lies in.
This 
upper bound limits the ability of OMP in producing a denser affinity graph, leading to a high risk of over-segmentation.

We incorporate the dropout technique in the previous section to address the connectivity issue in solving \eqref{eq:self-expression-omp} via OMP. Specifically, in Section~\ref{sec:Stochastic-SSC} we propose a flexible subspace clustering method that combines SSC with \eqref{eq:expectation-dropout-self-expression-sample-T}, and subsequently rewrite it as a consensus optimization problem.
Then, in Section~\ref{sec:alt-minimization-panelty-method} we present 
an efficient alternating minimization algorithm to solve the consensus problem.

\subsection{Stochastic Sparse Subspace Clustering}
\label{sec:Stochastic-SSC}

%
By combining the sample mean of the stochastic self-expressive model in \eqref{eq:expectation-dropout-self-expression-sample-T} and the sparsity constraint in \eqref{eq:self-expression-omp}, we propose a \emph{stochastic sparse subspace clustering} model as follows:
\begin{equation}
\begin{split}
& \!\min \limits_{\c_j} \frac{1}{T} \sum_{t=1}^T \| \x_j - \sum_i \xi_{i}^{(t)} c_{ij} \x_i \|_2^2  \\
& \st  \quad \|\c_j\|_0 \le s, ~~c_{jj}=0,
\end{split}
\label{eq:SSC-OMP-random-T-consensus}
\end{equation}
where $s$ controls the sparsity of the solution.\footnote{Due to the implicit squared $\ell_2$ regularization, the sparsity can be greater than the dimension of the subspace.}
Due to the stochastic nature of the dictionaries used in the $T$ subproblems and the sparsity constraint, we refer 
\eqref{eq:SSC-OMP-random-T-consensus} to Stochastic Sparse Subspace Clustering.

To understand the essence in solving problem~\eqref{eq:SSC-OMP-random-T-consensus}, we introduce $T$ auxiliary variables $\{\b_j^{(t)}\}_{t=1}^T$ and derive an equivalent formulation as follows:
\vspace{-5pt}
\begin{equation}
\begin{split}
&\!\!\!\!\!\!\!\! \min \limits_{\c_j, \{\b_j^{(t)}\}_{t=1}^T} \frac{1}{T} \sum_{t=1}^T \| \x_j - \sum_i \xi_{i}^{(t)} b^{(t)}_{ij} \x_i \|_2^2, \\
&\st ~~\b_j^{(1)}\!=\!\cdots\!=\!\b_j^{(T)}=\c_j, ~~ \|\b_j^{(t)}\|_0 \le s, \\
&~~~~~~~ b^{(t)}_{jj}=0,~~t=1,\cdots,T.
\label{eq:SSC-OMP-random-T-consensus-T-blocks}
\end{split}
\end{equation}
%
This is clearly a consensus problem over $T$ blocks.
Once the optimal solution $\c_j$ is found, we induce the affinity via $a_{ij} = \frac{1}{2}(|c_{ij}|+|c_{ji}|)$ and apply spectral clustering via normalized cut \cite{Shi-Malik:PAMI00} on this affinity matrix.
%
%
%
%

\myparagraph{Remark} In problem \eqref{eq:SSC-OMP-random-T-consensus-T-blocks}, the $T$ subproblems can be solved in parallel and each subproblem uses a small dictionary with $(1-\delta)N \ll N $ columns on average. This is appealing especially when the data is too large to fit into the memory.

\setlength{\textfloatsep}{8pt}
\begin{algorithm}[t]  
		\caption{\bf : Damped OMP}
		\label{alg:damped-OMP}
		\begin{algorithmic}[1]
            \REQUIRE Dictionary $\Xi$, $\II$, $\x_j \in \RR{^D}$, $\c_j$, $s$, $\lambda$ and $\epsilon$. 
            \STATE Initialize $k = 0$, residual $\q_j^{(0)} = \x_j$, and $S^{(0)} = \emptyset$. 
            \WHILE {$k < s$ and $\|\q_j^{(k)}\|_2 > \epsilon$}
            \STATE Find $i^\ast$ via \eqref{eq:damped-index-i-star} and update $S^{(k+1)} \leftarrow S^{(k)} \bigcup \{i^\ast \}$;
            \STATE Update $\b_j^{(k+1)}$ by solving \eqref{eq:S^3COMP-Penalty-update-b-j}; 
            \STATE Update $\q_j^{(k+1)} \leftarrow \x_j - \Xi \ \b_j^{(k+1)}$ and $k \leftarrow k+1$; 
			\ENDWHILE
            \ENSURE ${\b_j^\ast}$. 
		\end{algorithmic}
\end{algorithm}

\subsection{Consensus Orthogonal Matching Pursuit}
\label{sec:alt-minimization-panelty-method}

To efficiently solve problem \eqref{eq:SSC-OMP-random-T-consensus-T-blocks}, instead of 
solving problem \eqref{eq:SSC-OMP-random-T-consensus-T-blocks} exactly, we introduce a set of penalty terms and solve the relaxed problem as follows:
\vspace{-4pt}
\begin{equation}
\label{eq:S^3COMP-Penalty}
\begin{split}
&\!\!\!\!\min \limits_{\c_j, \{\b_j^{(t)}\}} \!\! \frac{1}{T} \sum_{t=1}^T \| \x_j \! -\! \sum_i \xi_{i}^{(t)} b^{(t)}_{ij} \x_i \|_2^2 + {\lambda \| {\b_j^{(t)}\! -\! {\c_j}} \|_2^2} \\
& ~\st ~~ \|\b_j^{(t)}\|_0 \le s, ~~b^{(t)}_{jj}=0, ~~t=1,\cdots,T,
\end{split}
\end{equation}
where $\lambda >0$ is a penalty parameter.
We solve problem \eqref{eq:S^3COMP-Penalty} by updating $\{\b_j^{(t)}\}_{j=1}^T$ and $\c_j$ alternately. 

\setlength{\textfloatsep}{8pt}
\begin{algorithm}[t]
		\caption{\bf Consensus OMP for Solving Problem \eqref{eq:S^3COMP-Penalty}}
		\label{alg:DOMP-Consensus-Panelty}
		\begin{algorithmic}[1]
            \REQUIRE $X = [\x_1, \dots, \x_N] \in \RR{^{D \times N}}$, $\x_j \in \RR{^D}$, parameters $s$, $\delta$, $\lambda$, $\epsilon$, $T$. 
            \STATE Sample $T$ subdictionaries $\left\{{\Xi^{(t)}}\right\}_{t=1}^T$ via \eqref{eq:xi-definition-Bernoulli}; 
            \WHILE {not converged}
            \STATE Given $\c_j$, solve $T$ subproblems for $\{\b_j^{(t)}\}_{t=1}^T$ in parallel via Algorithm \ref{alg:damped-OMP};
            \STATE Given $\{\b_j^{(t)}\}_{t=1}^T$, update $\c_j$ via $\c_j \leftarrow  \frac{1}{T}\sum_{t = 1}^T {\b_j^{(t)}}$; 
            \ENDWHILE
			\ENSURE $\c_j^\ast$.
		\end{algorithmic}
\end{algorithm}

\begin{enumerate}[leftmargin=*]
\vspace{-2pt}
\item \textbf{When $\c_j$ is fixed:} we solve for $\{\b_j^{(t)}\}_{j=1}^T$ in parallel from each of the $T$ subproblems as follows
\begin{equation}
\label{eq:S$^3$COMP-Penalty-$b-j-t$}
\begin{split}
&\min \limits_{\b_j^{(t)}} \|{\x_j} - \sum\limits_i {\xi_{i}^{(t)}} b_{ij}^{(t)}{\x_i}\|_2^2 + \lambda \| {\b_j^{(t)} - {\c_j}} \|_2^2, \\
& ~~~\st ~~ \|{\b^{(t)}_j} \|{_0} \le s, ~~b^{(t)}_{jj} = 0.
\end{split}
\end{equation}
%
%
%
%
%
Denote the index set for the preserved and dropped columns in the $t$-th subproblem 
with $\II^{(t)} := \{i: ~\xi^{(t)}_{i} > 0\}$ and $\J^{(t)} := \{i: ~\xi^{(t)}_{i} = 0\}$, respectively, and let $\Xi^{(t)}$ be the same as the data matrix $X$ except that columns indexed by 
$\J^{(t)}$ are set to zero vectors.
%
%
%
%
%
%
For clarity, 
we rewrite problem \eqref{eq:S$^3$COMP-Penalty-$b-j-t$} via dictionary $\Xi^{(t)}$ (but ignore the 
superscript $t$) as follows:
\vspace{-2pt}
\begin{equation}
\label{eq:S$^3$COMP-Penalty-b-j}
\begin{split}
&\min \limits_{\b_j } \|{\x_j} - \Xi\ \b_j \|_2^2 + \lambda \| {\b_j  - {\c_j}} \|_2^2, \\
& ~~~\st ~~ \|{\b_j} \|{_0} \le s, ~~b_{jj} = 0.
\end{split}
\end{equation}
To solve problem \eqref{eq:S$^3$COMP-Penalty-b-j} efficiently, we develop a greedy algorithm to update $\b_j$ from the support within the index set $\II$ of the 
preserved columns.\footnote{The reason to update $\b_j$ only from the support in $\II$ is to enlarge the support of the consensus solution $\c_j$ while keeping the efficiency. This is equivalent to use an enlarged sparsity parameter $s' > s$ in \eqref{eq:SSC-OMP-random-T-consensus} or \eqref{eq:SSC-OMP-random-T-consensus-T-blocks}.}
To be specific, we initialize the support set $S^{(0)}$ as an empty set, set the residual $\q_j^{(0)} = \x_j$,
and find the support set $S^{(k+1)}$ of the solution $\b_j$ by a greedy search procedure, 
\ie, incrementing $S^{(k)}$ by adding one index $i^\ast$ at each iteration via
\vspace{-4pt}
\begin{align}
\label{eq:damped-index-i-star}
i^\ast =\arg \max_{i \in \II \setminus S^{(k)} } \psi_i(\q_j^{(k)}, \c_j),
\vspace{-4pt}
\end{align}
%
where $\psi_i(\q_j^{(k)}, \c_j) =(\x_i^\top \q_j^{(k)})^2 + 2 \lambda \x_i^\top \q_j^{(k)}c_{ij} - \lambda c^2_{ij}$.
%
After updating $S^{(k+1)}$ via $S^{(k+1)} \leftarrow S^{(k)} \cup \{i^\ast\}$, we solve problem
\vspace{-2pt}
\begin{equation}
\label{eq:S^3COMP-Penalty-update-b-j}
\begin{split}
&~\min_{\b_j} {\| {\x_j - \Xi\ \b_j} \|^2_2}+\lambda \|\b_j - \c_j\|_2^2 \\
&~~~\st ~~ {\rm{supp}}( \b_j ) \subseteq {S^{(k+1)}}
\end{split}
\end{equation}
%
with a closed-form solution, and then compute the residual $\q_j^{(k+1)} = \x_j - \Xi \ \b_j^{(k+1)}$.
%

We summarize the steps for solving problem \eqref{eq:S$^3$COMP-Penalty-b-j} in Algorithm \ref{alg:damped-OMP}, termed as Damped Orthogonal Matching Pursuit (Damped OMP). 

\item \textbf{When $\{\b_j^{(t)}\}_{j=1}^T$ are fixed:} we solve $\c_j$ from problem
\begin{equation}
\label{eq:S^3COMP-Penalty-update-c-j}
\min \limits_{\c_j} \frac{\lambda}{T} \sum_{t=1}^T \| {\b_j^{(t)} - {\c_j}} \|_2^2,
\end{equation}
which has a closed-form solution $\c_j = \frac{1}{T}\sum_{t = 1}^T {\b_j^{(t)}}$.
\end{enumerate}

\begin{figure*}	
\centering
		{
			\includegraphics[clip=true,trim=1 0 0 0,width=0.525\columnwidth]{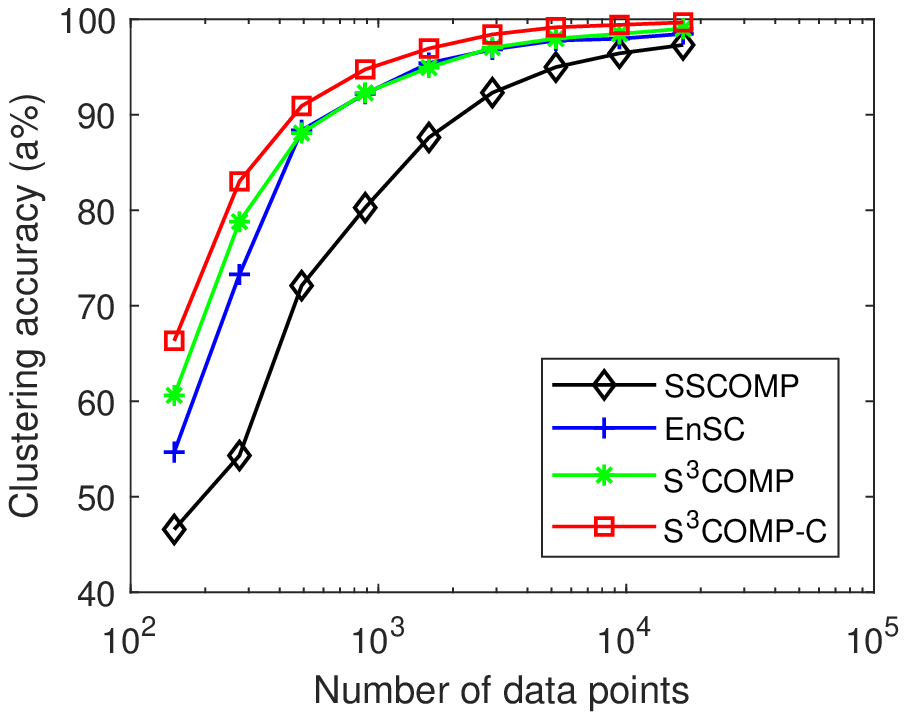}
			\label{fig:syn_allmethod_acc}
		}
		~
		{
			\includegraphics[clip=true,trim=1 0 0 0,width=0.525\columnwidth]{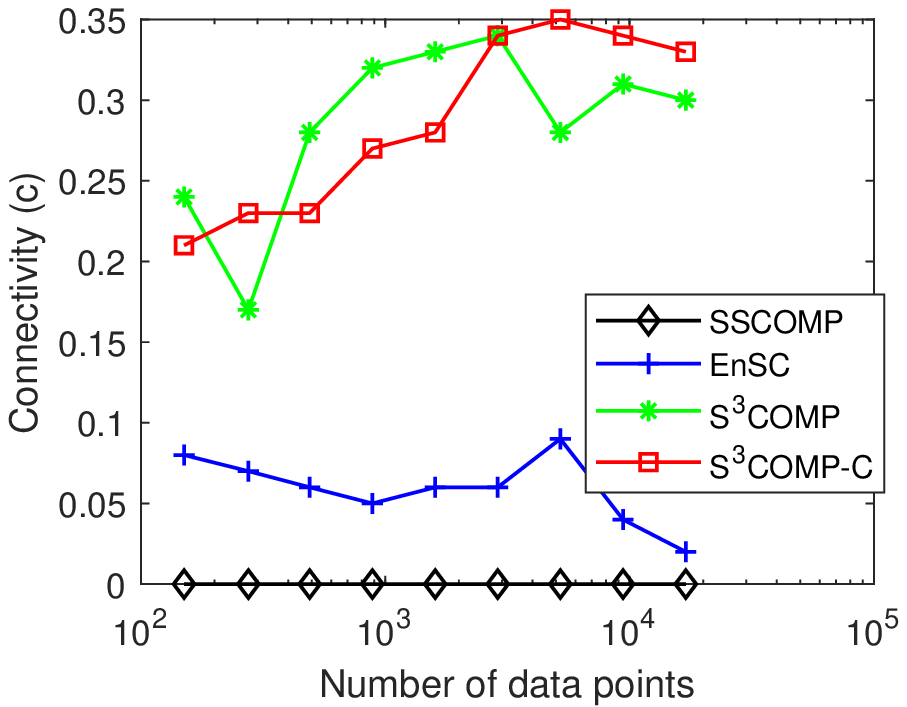}
			\label{fig:syn_allmethod_con}
		}
		~
		{
			\includegraphics[clip=true,trim=1 0 0 0,width=0.525\columnwidth]{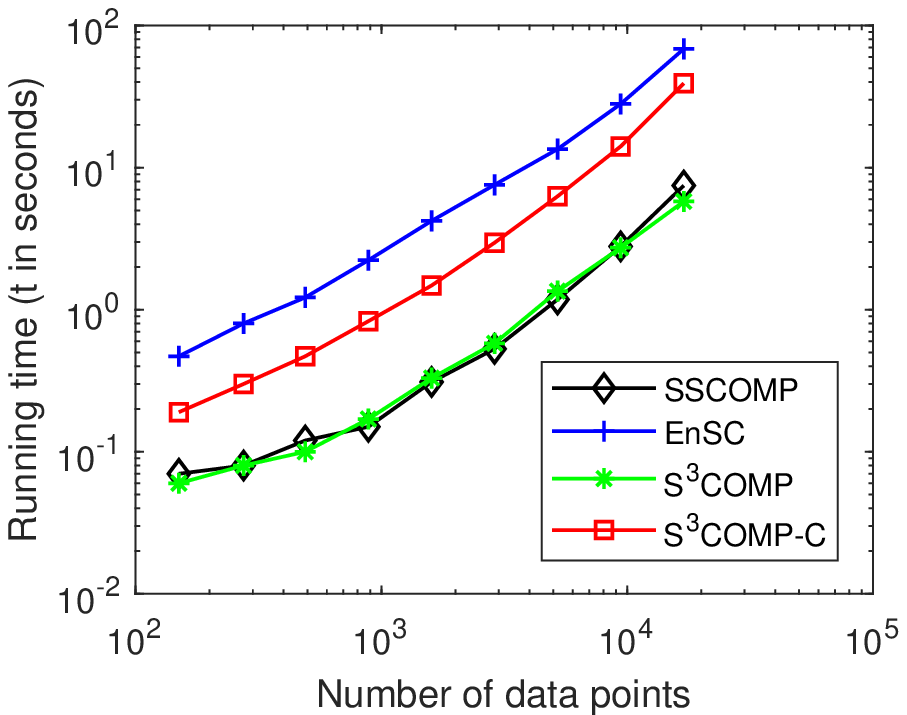}
			\label{fig:syn_allmethod_time}
		}
\caption{Performance comparison of {S}$^{3}$COMP-C, {S}$^{3}$COMP, EnSC and SSCOMP on synthetic data.}
\label{fig:syn_result_acc_con_time}		
\end{figure*}

We summarize the alternating minimization algorithm for solving the consensus problem~\eqref{eq:S^3COMP-Penalty} in Algorithm \ref{alg:DOMP-Consensus-Panelty}, termed Consensus OMP.
For clarity, we sort the whole procedure of our proposed subspace clustering approach in Algorithm \ref{alg:S$^3$COMP-C}, termed Stochastic Sparse Subspace Clustering via Orthogonal Matching Pursuit with Consensus (S$^3$COMP-C),
%
and we use S$^3$COMP to refer the approach that solves the consensus problem~\eqref{eq:S^3COMP-Penalty} via Algorithm \ref{alg:DOMP-Consensus-Panelty} only one outer iteration. Note that the support size of each solution $\b^{(t)}_j$ is 
up to $s$ and thus the support size of the solution ${\c_j}$ obtained via $\frac{1}{T}\sum_{t = 1}^T {\b_j^{(t)}}$ will be 
up to $sT$, 
leading to improved connectivity of the induced affinity graph.

%
%
%

\myparagraph{Convergence and Stopping Criterion} Similar to the convergence analysis in OMP~\cite{Pati:ASILOMAR93}, Algorithm~\ref{alg:damped-OMP} converges in at most $s$ steps.
%
%
For Algorithm \ref{alg:DOMP-Consensus-Panelty}, we stop it by checking whether the relative changes of $\c_j$ in two successive iterations is smaller than a threshold $\varepsilon$ or reaching the maximum iterations. 
%
Although we cannot prove the convergence of Algorithm \ref{alg:DOMP-Consensus-Panelty}, experiments on synthetic data and the real world data demonstrate a good convergence. In experiments, we observe that the number of the outer iterations is small, \ie, $T_0 = 3\sim5$ on real world datasets.

\setlength{\textfloatsep}{8pt}
\begin{algorithm}[t]
		\caption{\bf : S$^3$COMP-C}
		\label{alg:S$^3$COMP-C}
		\begin{algorithmic}[1]
            \REQUIRE $X = [\x_1, \dots, \x_N] \in \RR{^{D \times N}}$, $\x_j \in \RR{^D}$, parameters $s$, $\delta$, $\lambda$, $\epsilon$ and $T$. 
            \STATE Run Algorithm \ref{alg:DOMP-Consensus-Panelty};
            \STATE Define affinity via $a_{ij} = \frac{1}{2} (|c_{ij}| + |c_{ji}|)$;
            \STATE Run spectral clustering via normalized cut \cite{Shi-Malik:PAMI00};
            \ENSURE Segmentation matrix.
		\end{algorithmic}
\end{algorithm}

\myparagraph{Complexity Analysis}
In Algorithm~\ref{alg:DOMP-Consensus-Panelty}, it solves $T$ size-reduced subproblems via a damped OMP in parallel, and each subproblem requires
$N(1 - \delta)$ inner products. Thus, the computation complexity in this stage for each subproblem is $\mathcal{O}(D N^2(1-\delta)s)$ in one outer iteration.
%
The affinity matrix of S$^{3}$COMP and S$^{3}$COMP-C contains at most $s T N$ non-zero entries; whereas the affinity matrix of SSCOMP contains at most $sN$ nonzero entries. 
The eigenvalue decomposition of a sparse matrix 
using ARPACK requires $\mathcal{O}(s n T N)$ operations where $n$ is the number of subspaces (\ie, clusters).
%
%
%
%
%
%
%
%
%
%
%
%
%
While the affinity matrix of S$^{3}$COMP-C and S$^{3}$COMP may contain more nonzero entries (up to $sTN$), the affinity matrix is still sparse and thus the time complexity of eigenvalue decomposition in spectral clustering is $O(nsTN)$, which is slightly higher than $O(nsN)$ of SSCOMP.
For a data set of large size, we set $(1-\delta) \ll 1$ and solve $T$ size-reduced subproblems in parallel. This endorses S$^{3}$COMP-C and S$^{3}$COMP more flexible scalability. 

\section{Experiments}
\label{sec:experiments}


To evaluate the performance of our proposed approach, we conduct extensive experiments on both synthetic data and real world benchmark datasets.

\myparagraph{Methods and Metrics} 
We select eight state-of-the-art subspace clustering 
methods as baselines: SCC \cite{Chen:IJCV09}, LSR \cite{Lu:ECCV12}, LRSC \cite{Vidal:PRL14}, and several scalable subspace clustering methods, including SSCOMP \cite{You:CVPR16-SSCOMP}, EnSC \cite{You:CVPR16-EnSC}, OLRSC \cite{Shen:ICML16}, SR-SSC \cite{Abdolali:SP19}, and ESC \cite{You:ECCV18}. In experiments, we use the code provided by the authors for computing the self-expression matrix $C$ in which the parameter(s) is tuned to give the best clustering accuracy. For spectral clustering, we apply the normalized cut \cite{Shi-Malik:PAMI00} on the affinity matrix $A$ which is induced via $A = |C| + |C^\top|$, except for SCC, which has its own spectral clustering step.
The reported results in all the experiments of this section are averaged over 10 trials.
Following~\cite{You:CVPR16-SSCOMP}, we evaluate each algorithm with clustering accuracy\footnote{It is computed by finding the best alignment between the clustering index and the ground-truth labels under all possible permutations.} (acc:$a\%$),
subspace-preserving representation error (sre:$e\%$),
{connectivity\footnote{Let $\lambda_2^{(i)}$ be the second smallest eigenvalue of the normalized graph Laplacian corresponding to the $i$-th cluster \cite{Mohar:GTCA91}. The connectivity is computed by $c:=\min_i \{ \lambda_2^{(i)}\}_{i=1}^n$ for synthetic data. To show the improvement on average for real world data, we compute $\bar c:=\frac{1}{n}\sum_{i=1}^n  \lambda_2^{(i)}$.}  (conn:$c$), 
and running time\footnote{The running time of S$^3$COMP in Tables \ref{tbl:EYaleB-result} to \ref{tbl:GTSRB-result} is based on the maximum running time among $T$ subtasks plus the time of spectral clustering.} ($t$).
%
%
}

\begin{figure*}
\centering
		\subfigure[Clustering accuracy ($a\%$)]
		{
			\includegraphics[clip=true,trim=1 0 0 0,width=0.545\columnwidth]{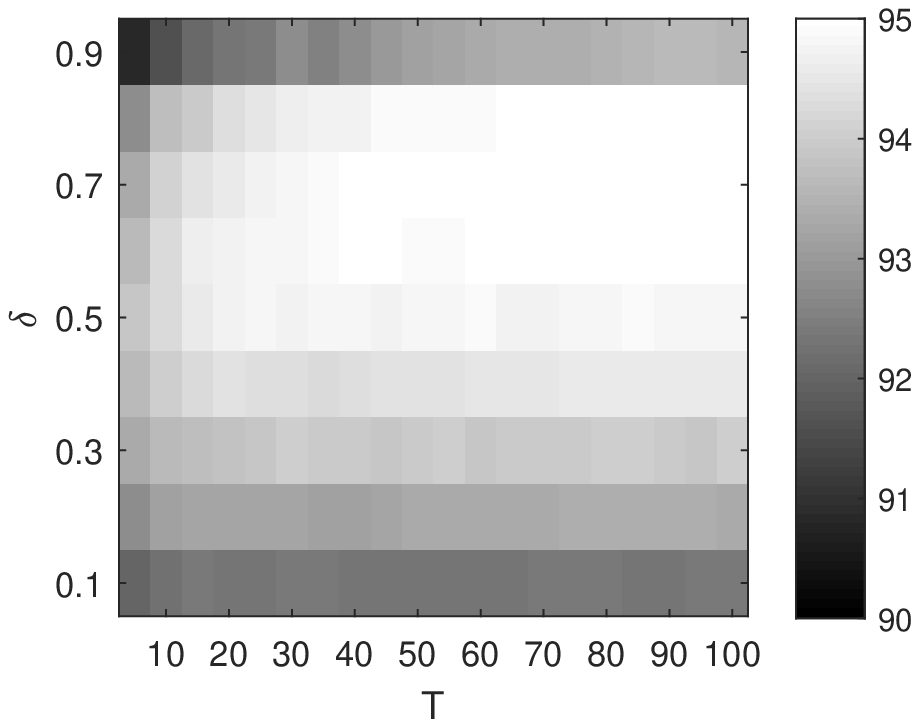}
			\label{fig:syn320_delta_T_acc}
		}
		~
		\subfigure[Connectivity ($c$)]
		{
			\includegraphics[clip=true,trim=1 0 0 0,width=0.545\columnwidth]{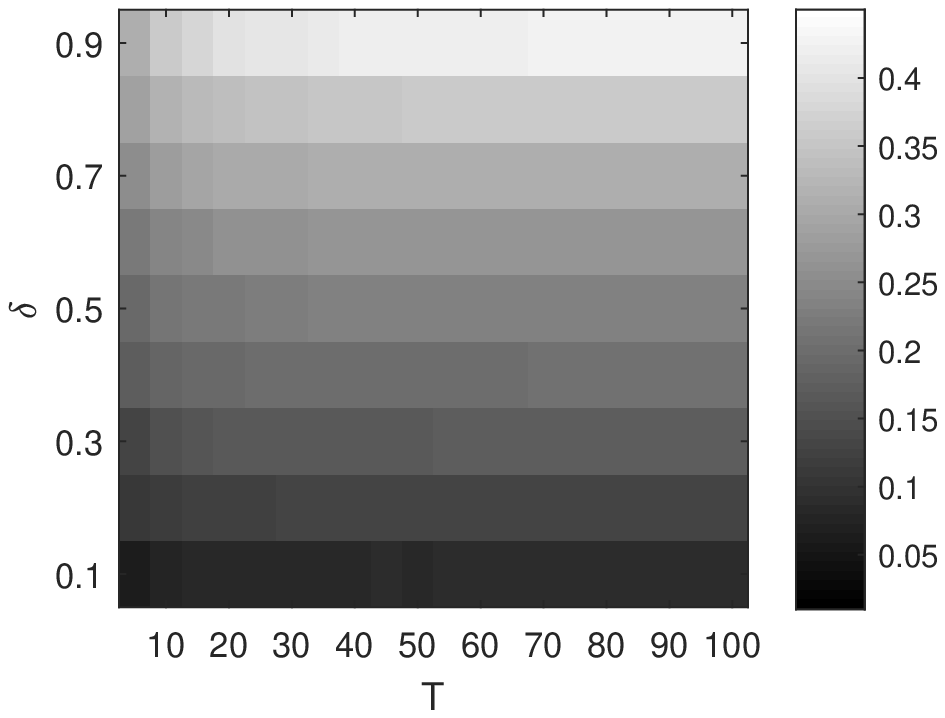}
			\label{fig:syn320_delta_T_con}
		}
		~
		\subfigure[Subspace-preserving rate ($1 - e\%$)]
		{
			\includegraphics[clip=true,trim=1 0 0 0,width=0.545\columnwidth]{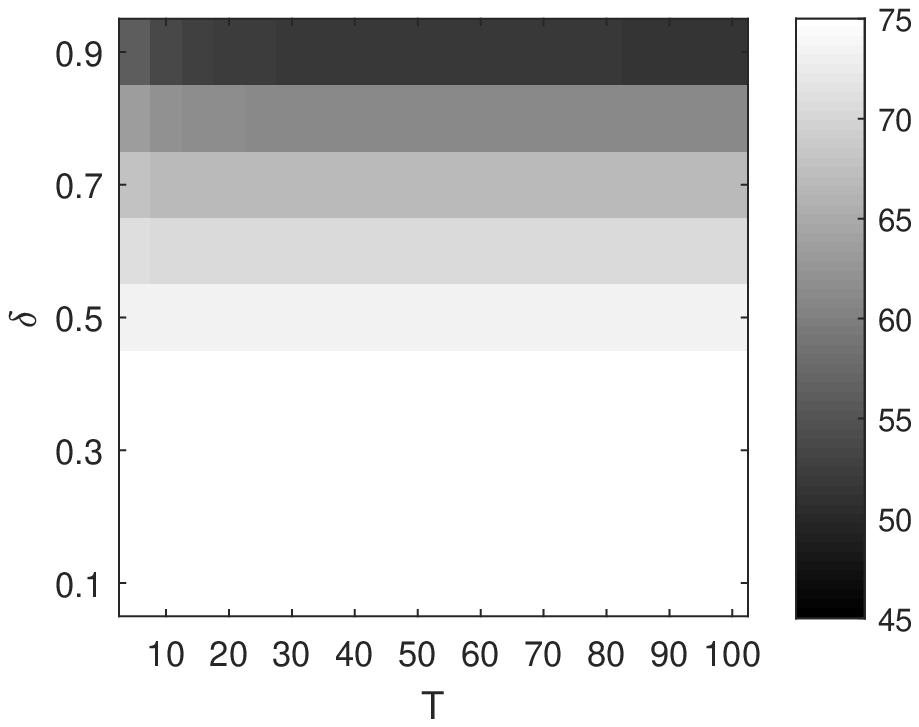}
			\label{fig:syn320_delta_T_sse}
		}
\caption{Performance of {S}$^{3}$COMP-C as functions of $T$ and $\delta$ on synthetic data of $N_i=320$. The intensity corresponds to the value.} 
\label{fig:syn320_delta_T-result}		
\vspace{-5pt}
\end{figure*}

\subsection{Experiments on Synthetic Data}
\label{sec:Synthetic Experiments}

\myparagraph{Setup} 
We follow the setting used in~\cite{You:CVPR16-SSCOMP} to randomly generate $n=5$ subspaces of dimension $d = 6$ in the ambient space $\RR^9$. Each subspace contains $N_i$ data points randomly sampled on a unit sphere of $\RR^9$, in which $N_i$ varies from $30$ to $3396$. Thus, the total number $N$ of data points varies from $150$ to $16980$. For a fair comparison, we use the same parameter $s=5$ as in~\cite{You:CVPR16-SSCOMP}. We set $T=15$ and select the dropout rate $\delta$ in $\{0.1,0.2,\cdots,0.9\}$.


We conduct experiments on the synthetic data with different data points per subspace and report the accuracy, connectivity, and subspace-preserving errors. 
We show each metric as a function of $N_i$, and present them as curves in Fig.~\ref{fig:syn_result_acc_con_time}. We observe that both S$^{3}$COMP-C and S$^{3}$COMP outperform SSCOMP in clustering accuracy and connectivity, especially when the density of data points is lower. It is clear that EnSC, S$^{3}$COMP and S$^{3}$COMP-C all improve the connectivity in all cases.
%
The computation time of S$^{3}$COMP is comparative to (or even lower than) SSCOMP. EnSC yields very competitive clustering accuracy to S$^{3}$COMP but the time cost is higher than S$^{3}$COMP-C.

To better understand the effects of parameters $\delta$ and $T$, we conduct experiments with S$^{3}$COMP-C on synthetic data of $N_i=320$ under varying $\delta \in \{0.1,\cdots, 0.9\}$ and $T \in \{5,10,\cdots,100\}$. 
The performance of each metric is recorded as a function of $\delta$ and $T$, and displayed as intensity of a gray image in Fig.~\ref{fig:syn320_delta_T-result}. We observe that the clustering accuracy tends to being stable even using a high dropout rate (\eg, $\delta=0.85$) whenever $T$ is large than 10.
Roughly speaking, higher dropout rate leads to higher connectivity and more efficient algorithm.
%
Thought we also observe that using a higher dropout rate leads to slightly higher subspace-preserving errors\footnote{This does not contribute to improve the algebraic connectivity~\cite{Mohar:GTCA91}. Thus, the exact relation of the algebraic connectivity with respect to $\delta$ is not simply monotonous.},
it does not necessarily degenerate the clustering accuracy. This is because that the improved connectivity could not only help to avoid over-segmenting the data points in same subspaces but also make the connected data points within the same subspaces have more compact clusters in spectral embedding.

\subsection{Experiments on Real World Datasets}
\label{sec:Real Experiments}

In this subsection, we demonstrate the performance of the proposed method on four benchmark datasets, including Extended Yale B (EYaleB) \cite{Kriegman:TPAMI01}, Columbia Object Image Library (COIL100) \cite{Nene:TR96}, MNIST~\cite{LeCun:PIEEE1998}, and German Traffic Sign Recognition Benchmark (GTSRB) \cite{STALLKAMP:NN12}.

\myparagraph{Dataset Descriptions}
\text{Extended Yale B} 
contains 2432 frontal facial images of $38$ individuals under $64$ different illumination conditions, each of size $192 \times 168$. In our experiments, we use the images of all the 38 individuals and resize each image into $48 \times 42$ pixels and concatenate the raw pixel in each image as a 2016-dimensional vector.

\text{COIL100} 
contains 7,200 gray-scale images of 100 different objects. Each object has 72 images taken at pose intervals of 5 degrees. We resize each image to the size $32 \times 32$, and concatenate the gray-pixels in each image as a 1024-dimensional vector.

\text{MNIST} 
contains 70,000 grey-scale images of handwritten digits $0-9$. 
In addition to the whole dataset (denoted MNIST70000), we also prepared two subsets---MNIST4000 and MNIST10000, which are generated by random sampling $N_i=400$ and $N_i=1000$ images per category, respectively. 
For each image, we compute a feature vector of dimension 3,472 using the scattering convolution network \cite{Bruna:PAMI13} and then reduce the dimension to $500$ using PCA.

\text{GTSRB} 
contains 43 categories of street sign images with over 50,000 samples in total. We preprocess the dataset as in ESC \cite{You:ECCV18}, which results in an imbalanced dataset of 12,390 images in 14 categories. Each image is represented by a 1568-dimensional HOG feature provided with the database. The feature vectors are mean-subtracted and projected to dimension 500 by PCA.

\myparagraph{Setup} Note that all feature vectors are normalized to have unit $\ell_2$ norm before performing subspace clustering. For a fair comparison, we set $s=10$ for MNIST and $s=5$ for Extended Yale B, respectively, as in SSCOMP \cite{You:CVPR16-SSCOMP}, and set $s=3$ for GTSRB and COIL100.\footnote{In practice, the parameter $s$ is set to be equal to (or slightly less than) the intrinsic dimension of the data, which could be estimated.} For the experiments on the real world datasets, we set $T=15$ and select the dropout rate $\delta$ in $\{0.10,0.20,\cdots,0.90\}$.

\begin{table}[h]
\small
\centering
\begin{threeparttable}
\resizebox{0.45\textwidth}{!}{
\begin{tabular}{l|c c c c }
\hline
\multirow{2}{*}{Method} &\multicolumn{4}{c}{Extended Yale B}  \\
                        & acc (a\%) & sre (e\%) &conn ($\bar c$) &t (sec.)  \\
\hline
SCC    &12.80 &- &- &615.69  \\
OLRSC  &26.84 &95.98 &\bf{0.6284} &98.25  \\
LSR    &63.99 &87.57 &\underline{0.5067} &3.21 \\
LRSC   &63.17 &88.75 &0.4526 &7.20   \\
EnSC   &61.20 &23.14&0.0550 &52.98  \\
SR-SSC &62.11 &- &- &79.46  \\
SSCOMP &77.59 &\bf{20.13} &0.0381 &\underline{2.54}  \\
ESC$^\ast$    &\bf{87.58} &-& -&28.01   \\
\hline
{S}$^{3}$COMP   & 81.61 &\underline{20.18} &0.0723 &\bf{1.92}   \\ 
{S}$^{3}$COMP-C &\underline{87.41} &{20.28} &0.0667 &5.05  \\ 
\hline
\end{tabular}
}
\\[3pt]
\caption{Performance comparison on EYaleB where `-' denotes the metric cannot be computed properly. ESC$^\ast$ uses different way to define affinity from the self-expression coefficients.}
\label{tbl:EYaleB-result}
\end{threeparttable}
\end{table}

\begin{table}[h]
\small
\centering
\begin{threeparttable}
\resizebox{0.45\textwidth}{!}{
\begin{tabular}{l|c c c c }
\hline
\multirow{2}{*}{Method} &\multicolumn{4}{c}{COIL100}  \\
                        & acc (a\%) & sre (e\%) &conn ($\bar c$) &t(sec.)  \\
\hline
SCC  &55.24 &- &- &479.13  \\
%
%
LRSC &50.10 &96.43 &\bf{0.7072} &25.11  \\
LSR   &48.22 &94.95 &\underline{0.5246} &62.91  \\
SSCOMP &49.88 &14.03 &0.0060 &\underline{13.33}  \\ 
ESC   &56.90 &-& -&56.31   \\
SR-SSC    &58.85 &- &- &204.38  \\
EnSC   &{63.94} &{4.36} &0.0163 &19.03  \\
\hline
{S}$^{3}$COMP   &\underline{71.47} &\underline{3.35} &0.0081 &\bf{7.68}   \\ 
{S}$^{3}$COMP-C &\bf{78.89} &\bf{3.15} &0.0077 &20.10  \\ 
\hline
\end{tabular}
}
\\[3pt]
\caption{Performance comparison on COIL100 where `-' denotes the metric cannot be computed properly.} 
\label{tbl:COIL100-result}
\end{threeparttable}
\end{table}

\begin{table*}[tb]
\centering
\small
\begin{threeparttable}
\resizebox{0.75\textwidth}{!}{
\begin{tabular}{l|c c c c |c c c c }
\hline
\multirow{2}{*}{Method} &\multicolumn{4}{c|}{MNIST4000} &\multicolumn{4}{c}{MNIST10000} \\
                        & acc (a\%) & sre (e\%) &conn ($\bar c$) &t (sec.)  & acc (a\%) & sre (e\%) &conn ($\bar c$) &t (sec.) \\
\hline
LSR     &80.02 &78.53 &0.6075 &14.79 &81.75 &80.22 &0.6389 &147.98  \\
LRSC    &85.61 &79.87 &\underline{0.6419} &4.77 &89.60 &81.36 &\underline{0.6646} &12.87   \\
SCC     &71.30 &- &- &70.75 &72.20 &- &- &218.16 \\
OLRSC   &65.32 &85.70 &\bf{0.8660} &47.4  &67.62 &86.11 &\bf{0.8738} &217.43  \\
ESC     &87.22 &-& -&27.98 &90.76 &-&-&59.41  \\
EnSC    &85.85 &\bf{20.40}&0.1117 &35.89 &85.94& \bf{16.63}&0.0938&89.21  \\
SSCOMP  &91.14 &34.26 &0.1371 &\bf{3.63} & 93.80&{32.08} &0.1212 &\underline{11.99}  \\
SR-SSC  &91.70 &- &- &39.24 &90.05&-&-&79.87  \\
\hline
{S}$^{3}$COMP & \bf{94.30}&\underline{33.15} &0.1529 &\underline{4.70} &\underline{95.73} &\underline{30.11} &0.1720 &\bf{9.14}  \\ 
{S}$^{3}$COMP-C &\underline{94.27} &{33.26} &0.1527 &12.88 &\bf{95.74} &33.15 &0.1719 &26.50  \\ 
\hline
\end{tabular}
}
\\[3pt]
\caption{ Performance comparison on MNIST where `-' denotes the metric cannot be computed properly.}
\label{tbl:MNIST-result}
\end{threeparttable}
\vspace{-8pt}
\end{table*}



\myparagraph{Results}
The results on {Extended Yale B} are listed in Table \ref{tbl:EYaleB-result}. We can read that S$^{3}$COMP-C and S$^{3}$COMP improve the clustering accuracy roughly $10\%$ and $4\%$ over SSCOMP, respectively, and S$^{3}$COMP-C yields the second best clustering accuracy. The connectivity is improved while keeping a comparable or even lower subspace-preserving errors and computation cost. While ESC yields the best clustering accuracy, the time cost is much heavier. LSR, LRSC and OLRSC have good connectivity, but the subspace-preserving errors are worse and thus the accuracy is around $60\%$. While EnSC also has a good connectivity and a low subspace-preserving error, the accuracy and computation time are inferior to S$^{3}$COMP-C and S$^{3}$COMP.

In Table \ref{tbl:COIL100-result}, we report the results on {COIL100}. We can read that S$^{3}$COMP-C and S$^{3}$COMP yield the leading clustering accuracy and keeping the low subspace-preserving errors.
EnSC yields the third best clustering accuracy and subspace-preserving error, and keeps a better connectivity, due to taking a good tradeoff  between the $\ell_1$ and the $\ell_2$ norms. Note that the best three methods S$^{3}$COMP-C, S$^{3}$COMP and EnSC all yield very low subspace-preserving error and they share an (implicit or explicit) $\ell_2$ norm. 

\begin{table}[h]
\small
\centering
\begin{threeparttable}
\resizebox{0.45\textwidth}{!}{
\begin{tabular}{l|c c c c }
\hline
\multirow{2}{*}{Method} &\multicolumn{4}{c}{MNIST70000}  \\
                        & acc (a\%) & sre (e\%) &conn ($\bar c$) &t (sec.)  \\
\hline
OLRSC  &M &- &- &- \\
SR-SSC     &87.22&-&-&585.31  \\
SSCOMP$^\dag$  &81.59 &\underline{28.57} &0.0830 &\underline{280.58} \\
ESC    &90.87 &-&-&596.56  \\
EnSC    &93.67& \bf{15.30} &0.0911&932.89  \\
\hline
%
{S}$^{3}$COMP$^\dag$  &\underline{96.31} &30.12 &\bf{0.1569} &\bf{218.72}  \\ 
%
{S}$^{3}$COMP-C$^\dag$ &\bf{96.32} &30.11 &\bf{0.1569} &{416.84}  \\
\hline
\end{tabular}
}
\\[3pt]
\caption{Performance comparison on MNIST where `-' denotes the metric cannot be computed properly, 'M' means that the memory limit of 64G is exceeded. $^\dag$: The ending eleven eigenvectors associating with the smallest eleven eigenvalues are used in spectral clustering and the details are provided in the supporting material.}
\label{tbl:MNISTall-result}
\end{threeparttable}
\end{table}

The experiments on {MNIST} are provided in Table \ref{tbl:MNIST-result} and \ref{tbl:MNISTall-result}. Again, we can observe that S$^{3}$COMP-C still improves the clustering accuracy around $2\sim3\%$ on MNIST4000 and MNIST10000 
with improved connectivity than SSCOMP and keeping comparable subspace-preserving errors. On MNIST70000, SSCOMP yields seriously degenerated result than S$^{3}$COMP-C, S$^{3}$COMP and EnSC, due to the connectivity issue.
While EnSC has the lowest subspace-preserving error, the connectivity and the time cost are not in a good tradeoff. 
Note that LSR, LRSC, SCC and OLRSC cannot get results because of the memory limit of 64G; whereas S$^{3}$COMP-C and S$^{3}$COMP inherit the computation efficiency of SSCOMP.

\begin{table}[h]
\centering
\small
\begin{threeparttable}
\resizebox{0.45\textwidth}{!}{
\begin{tabular}{l|c c c c }
\hline
\multirow{2}{*}{Method} &\multicolumn{4}{c}{GTSRB}  \\
                        & acc (a\%) & sre (e\%) &conn ($\bar c$) &t (sec.)  \\
\hline
LSR   &73.93 &82.80 &0.6185 &290.97  \\
LRSC &87.28 &78.97 &\underline{0.6367} &15.85  \\
SCC &70.82 &- &- &237.01  \\
OLRSC &82.42 &77.15 &\bf{0.7606} &291.38   \\
SR-SSC    &78.42 &- &- &223.34  \\
SSCOMP &82.52 &5.42 &0.0213 &15.43  \\
EnSC   &86.05 &\bf{0.81}&0.0095 &33.46 \\
ESC   &90.16 &-& -&32.13   \\
\hline
{S}$^{3}$COMP & \underline{95.25}&\underline{2.40} &0.0576 &\bf{3.13}   \\ 
{S}$^{3}$COMP-C  &\bf{95.54} &{2.41} &0.0573 &\underline{7.10}  \\ 
\hline
\end{tabular}
}
\\[3pt]
\caption{Performance comparison on GTSRB where `-' denotes the metric cannot be computed properly.}
\label{tbl:GTSRB-result}
\end{threeparttable}
\end{table}

In Table \ref{tbl:GTSRB-result}, we show the results on {GTSRB}. While GTSRB is an imbalanced dataset, surprisingly, we can again observe that the proposed S$^{3}$COMP and S$^{3}$COMP-C outperform the listed baseline algorithms and achieve satisfactory results in all four metrics. 
For EnSC, while it yields the lowest subspace-preserving error, the low connectivity leads to inferior clustering result. Due to the imbalance in data distribution, it is hard to find a good tradeoff between the $\ell_1$ and $\ell_2$ norms.


\subsection{More Evaluations}
\label{sec:further-evaluations}

\begin{figure}[htb]
\centering
\vspace{-2pt}
\subfigure[Synthetic Data]
{
    \includegraphics[width=0.45\columnwidth]{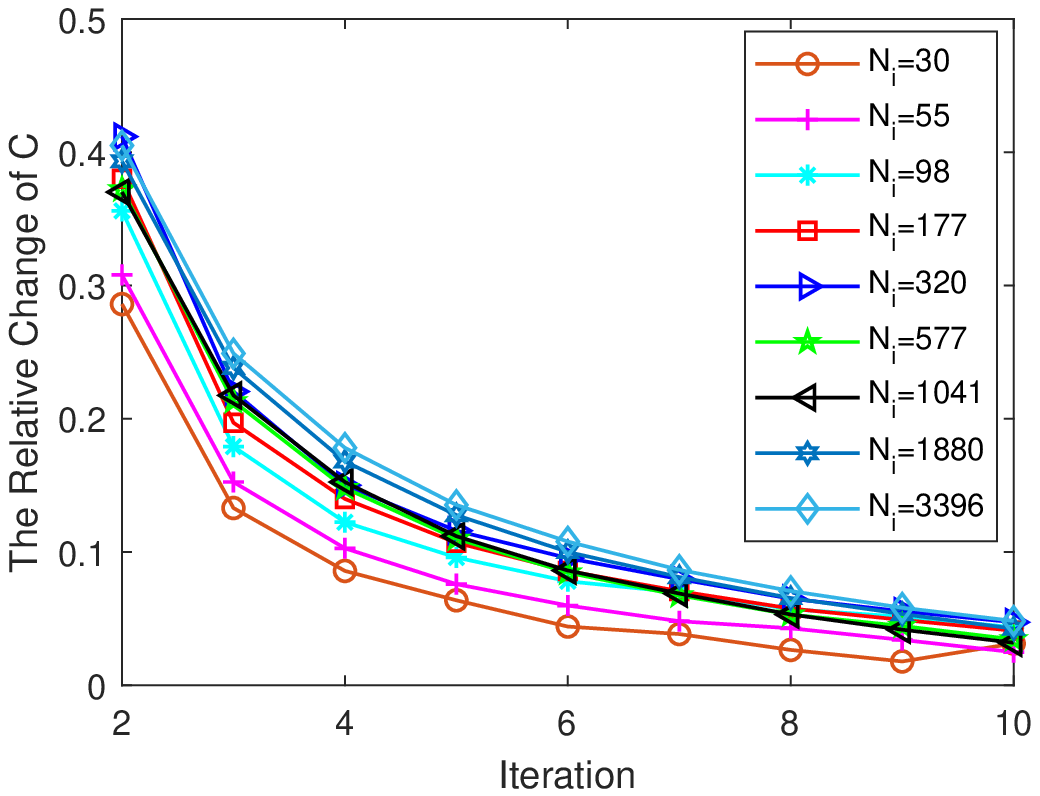}
    \label{figure:sub1-C}
}
\hspace{-3.5mm}
\subfigure[Real World Data]
{
    \includegraphics[width=0.45\columnwidth]{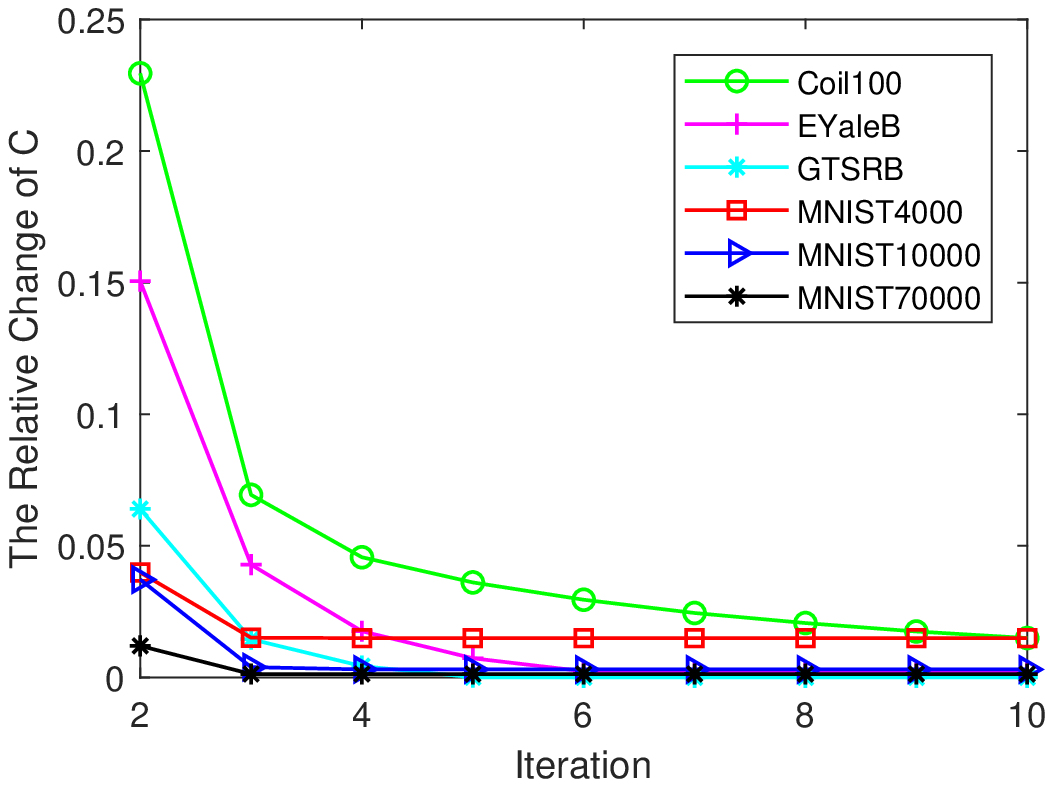}
}
\caption{The relative changes of $C$ in successive outer iterations.}
\vspace{-2pt}
\label{fig:C-Converge-Iter}
\end{figure}

\myparagraph{Convergence Behavior}
To evaluate the convergence of the proposed S$^{3}$COMP-C, we show the relative change of the self-expression matrix $C$ in two successive iterations on synthetic data and real world datasets in Fig. \ref{fig:C-Converge-Iter}. 
We observe that the self-expression matrix becomes stable after a few iterations. This confirms the convergence of Algorithm \ref{alg:DOMP-Consensus-Panelty}.

\begin{figure}[h]
\vspace{-3pt}
\centering
    \includegraphics[width=0.75\columnwidth]{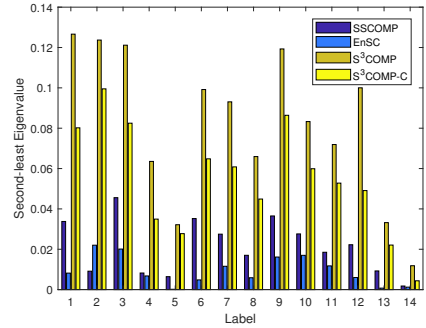}
\caption{Histograms of second minor eigenvalues of the normalized graph Laplacian for each category of GTSRB.}
\label{fig:GTSRB_conn_update}
\vspace{-2pt}
\end{figure}

\myparagraph{Improvements in Connectivity} To better observe the connectivity improvements of the proposed approach, we display the histogram of the second smallest eigenvalues of the normalized graph Laplacian corresponding to each category of GTSRB in Fig. \ref{fig:GTSRB_conn_update}. Note that the second minor eigenvalue of a normalized graph Laplacian with respect to each category measures the \emph{algebraic connectivity} \cite{Mohar:GTCA91}. The dramatic improvements in the second minor eigenvalues intuitively indicate significant improvements in connectivity.

\begin{figure}	
\vspace{-3pt}
\centering
		\subfigure[]
		{
			\includegraphics[clip=true,trim=2 0 5 5, width=0.455\columnwidth]{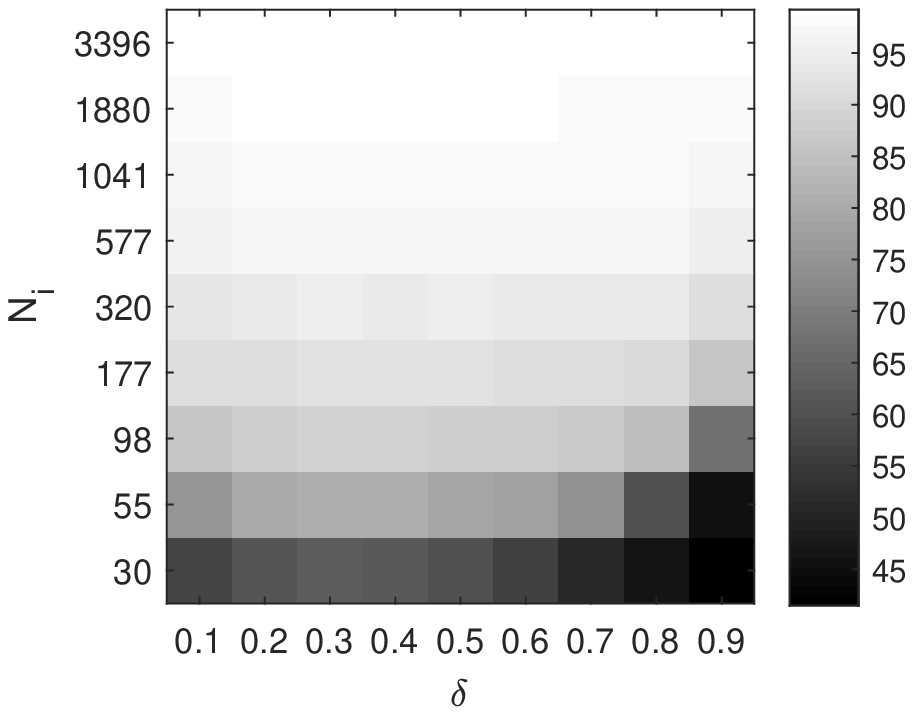}
			\label{fig:syn_consensus_delta_acc}
		}
		\subfigure[]
		{
			\includegraphics[clip=true,trim=2 0 5 5, width=0.485\columnwidth]{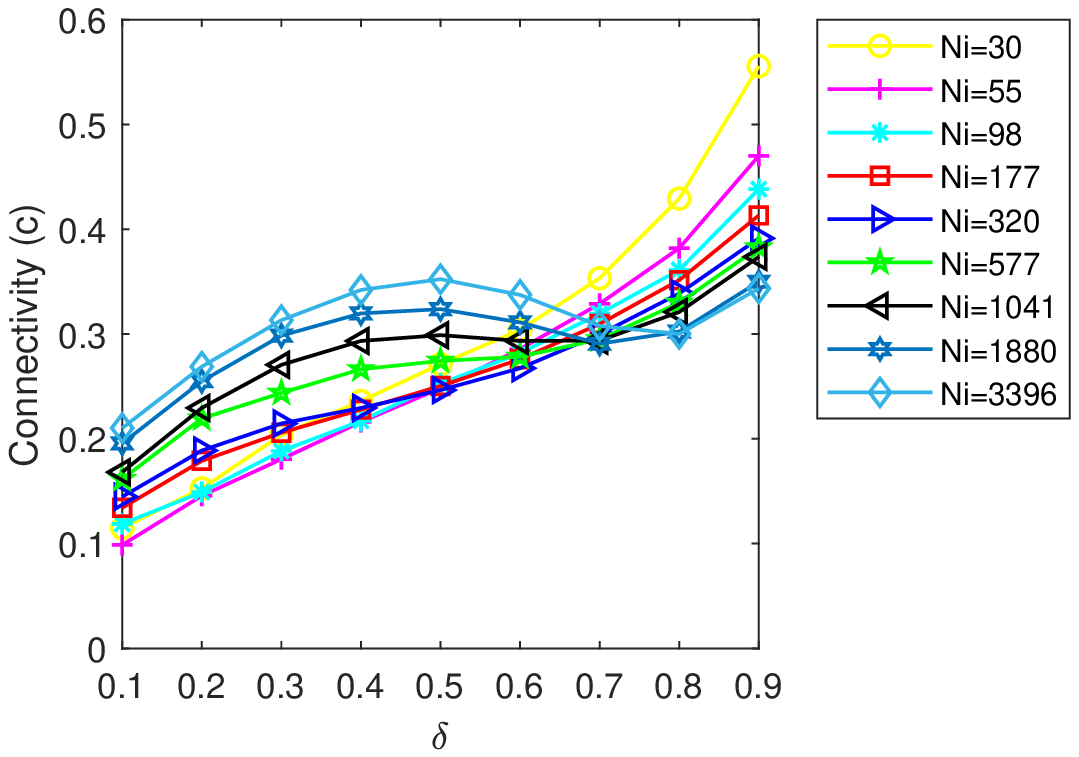}
			\label{fig:syn_consensus_delta_con}
		}
\caption{Evaluation on effect of dropout rate $\delta$ in S$^3$COMP-C on synthetic data of different number of data points per subspace. (a) Clustering accuracy (a\%) as a function of $\delta$ and $N_i$. (b) Connectivity $c$ as a function $\delta$. } 
\label{fig:syn-result-acc-con-sre-vs-delta-Ni}		
\vspace{-2pt}
\end{figure}

\myparagraph{Evaluation on Dropout Rate $\delta$} 
%
To evaluate the effect of varying the dropout rate $\delta$, we record the performance of {S}$^{3}$COMP-C using different dropout rate on synthetic data sets with different number of data points per subspace. Experimental results are presented in Fig.~\ref{fig:syn-result-acc-con-sre-vs-delta-Ni}. We observe that when the density of the data points increases, the clustering accuracy remains relatively stable when increasing the dropout rate.
Thus when the density of data points is higher we can use a larger dropout rate to discard more data points. This confirms that the dropout strategy actually leads to a flexible scalability, while building a desirable tradeoff between the computation efficiency and clustering accuracy.
%




\section{Conclusion}
\label{sec:conclusion}

We introduced a dropout strategy in the self-expressive model for subspace clustering.
By using Bernoulli random variables, we proved that the dropout in self-expressive model is equivalent to add a squared $\ell_2$ norm regularization.
Moreover, we 
proposed a scalable and flexible subspace clustering approach, which is formulated as a consensus optimization problem.
We solved the consensus problem by an alternating minimization algorithm which consists of a set of damped orthogonal matching pursuits and an average operation.
This leads to a principled and flexible way to improve the connectivity of the induced affinity graph and achieves a desirable tradeoff between the computation efficiency and clustering accuracy.
Extensive experiments on synthetic data and real world data have validated the efficiency and the effectiveness of our proposal.
%

\section*{Acknowledgment} This work is supported by the National Natural Science Foundation of China under grant 61876022, and the Open Project Fund from the MOE Key Laboratory of Machine Perception, Peking University. 

%

%
%
%
%
%

%

\appendices
\mbox{}

The appendices are organized as follows. In Section \ref{sec:append-proof}, we provide the proof for Theorem \ref{lemma:dropout-LS}. In Section \ref{sec:append-OMP-DOMP}, we present the review of the OMP algorithm and the derivation of the Damped OMP algorithm. In Section \ref{sec:append-experiments-more}, we provide details for the evaluation metrics and present additional evaluation results.

%

\section{Proof of Theorem~\ref{lemma:dropout-LS}}
\label{sec:append-proof}

\begin{proof}
Let $\{ \xi_i \}_{i=1}^N$ be i.i.d. random variables with distribution specified in \eqref{eq:xi-definition-Bernoulli}. We have
$\EE[\xi_{i}] =1$, $\EE [\xi_i\xi_j] = \EE[\xi_i] \EE [\xi_j] = 1$ if $i \neq j$, and $\EE [\xi_{i}^2] = \frac{1}{1-\delta}$.
Furthermore, we note that
\begin{equation}
\begin{split}
&\EE \| \x_j - \sum_i \xi_i c_i \x_i \|_2^2\\
=& \| \x_j \|_2^2 - 2 \EE \sum_i \xi_i c_i \x_j^\top \x_i + \EE \| \sum_i \xi_i \c_i \x_i\|_2^2,
\end{split}
\end{equation}
where the second term $\EE \sum_i \xi_i c_i \x_j^\top \x_i = \sum_i c_i \x_j^\top \x_i$. For the third term, we have
\begin{equation}
\begin{split}
 &\EE \| \sum_i \xi_i \c_i \x_i\|_2^2 \\
=& \sum_i \EE \| \xi_i c_i \x_i \|_2^2 + 2 \EE \sum_{i,k:i\neq k} c_i c_k \x_i^\top \x_k \xi_i \xi_k \\
=& \sum_i \EE [\xi_i^2] \| c_i \x_i \|_2^2 + 2 \sum_{i,k: i \neq k} c_i c_k \x_i^\top \x_k \EE [\xi_i \xi_k] \\
=& \frac{1}{1-\delta} \sum_i \| c_i \x_i \|_2^2 + 2 \sum_{i,k: i \neq k} c_i c_k \x_i^\top \x_k \\
=& \| \sum_i c_i \x_i \|_2^2 + (\frac{1}{1-\delta}-1) \sum_i \| c_i \x_i \|_2^2 \\
=& \| \sum_i c_i \x_i \|_2^2 + (\frac{\delta}{1-\delta}) \sum_i \| \x_i \|_2^2 c_i^2.
\end{split}
\end{equation}
Therefore, we have:
\begin{equation}
\begin{split}
&\EE \| \x_j - \sum_i \xi_i c_i \x_i \|_2^2 \\
=& \| \x_j \|_2^2 - 2 \sum_i c_i \x_j^\top \x_i + \| \sum_i \c_i \x_i\|_2^2 + \frac{\delta}{1-\delta} \sum_i \|\x_i\|_2^2 \c_i^2 \\
=& \| \x_j - \sum_i c_i \x_i \|_2^2 + \frac{\delta}{1-\delta} \sum_i \|\x_i\|_2^2 \c_i^2.
\end{split}
\end{equation}
This completes the proof.

\end{proof}

\section{Derivation of Damped Orthogonal Matching Pursuit}
\label{sec:append-OMP-DOMP}

\subsection{Review of OMP \cite{Pati:ASILOMAR93}}

Consider $\x_j \in \RR{^D}$. OMP is a greedy procedure for solving the following optimization problem
\begin{equation}
\min \limits_{\b_j} \left\| \x_j -X \b_j \right\|_2^2 \quad \text{s.t.} \quad \|\b_j\|_0 \le s, ~~b_{jj}=0.
\label{eq:SSC-OMP}
\end{equation}
Specifically, OMP keeps a working set $S^{(k)} \subseteq \{1, \cdots, N\}$, initialized as $S^{(0)} = \emptyset$ and incremented by one element at each iteration, and a current solution $\b_j^{(k)}$ that is supported on $S^{(k)}$.
Thus, we have that the set $S^{(k)}$ contains at most $k$ elements for each $k$. At the $k$-th iteration, $S^{(k)}$ and $\b_j^{(k)}$ are updated to $S^{(k+1)}$ and $\b_j^{(k+1)}$, respectively.
We consider the following optimization problem
\begin{equation}
\label{eq:SSC-OMP-local}
\begin{split}
    &  \min \limits_{\b_j} \|\x_j - X \b_j\|_2^2 \\
    &\st \quad \|\b_j\|_0 \le k + 1,  [\b_j]_{S^{(k)}} = [\b_j^{(k)}]_{S^{(k)}}.
\end{split}
\end{equation}
%
To update $S^{(k)}$ to $S^{(k+1)}$, one additional nonzero entry whose position and value is chosen so that the objective function in \eqref{eq:SSC-OMP-local} is minimized. We may rewrite this optimization into an equivalent form
\begin{equation}
\min\limits_{b_{ij} \in \RR} \min\limits_{i\in\{1, \cdots, N\}} \|\q_j^{(k)} - \x_i b_{ij}\|_2^2,
\end{equation}
where $\q_j^{(k)}:= \x_j - X \b_j^{(k)}$.
It is easy to see that the optimal $i^\ast$ is such that it maximizes 
$|\q_j^\top \frac{\x_i}{\|\x_i\|_2} |$, since that
\begin{equation}
\begin{split}
&\|\q_j^{(k)} - \x_i b_{ij}\|_2^2 \\
=& \|\q_j^{(k)}\|^2_2 + \|\x_i\|^2_2 b^2_{ij} - 2 \q_j^{(k)\top} \x_i b_{ij} \\
=& \|\q_j^{(k)}\|^2_2 + \|\x_i\|^2_2 (b_{ij} -   \frac{\q_j^{(k)\top} \x_i }{\|\x_i\|^2_2})^2 - \frac{(\q_j^{(k)\top} \x_i)^2}{\|\x_i\|^2_2} \\
=& \|\q_j^{(k)}\|^2_2 + (b_{ij} -   \q_j^{(k)\top} \x_i )^2 - (\q_j^{(k)\top} \x_i)^2,
\end{split}
\end{equation}
where the last equality comes from the fact that $\x_i$ is of unit $\ell_2$ norm, \ie, $\| \x_i \|_2^2 =1$. We may therefore set $S^{(k+1)} = S^{(k)} \cup \{i^*\}$. 
Then, OMP solves an optimization problem as follows:
\begin{equation}
\begin{split}
&\b_j^{(k+1)} = \mathop {\arg \min }\limits_{\b_j} \|\x_j - X \b_j\|_2^2 \\
&\st \quad \text{supp}(\b_j) \subseteq S^{(k+1)}.
\end{split}
\end{equation}
%
Instead of freezing the entries of $\b_j^{(k+1)}$ that are in the support set of the previous iteration $S^{(k)}$, OMP also optimizes for the values of those entries supported on $S^{(k)}$.

\subsection{Derivation of Damped OMP}

Consider the following optimization problem:
\begin{equation}
\begin{split}
& \min \limits_{\b_j} \left\| \x_j -X \b_j \right\|_2^2 + \lambda \left\| \b_j - \c_j \right\|_2^2 \\
&~~ \text{s.t.} \quad \|\b_j\|_0 \le s, ~~b_{jj}=0.
\end{split}
\label{eq:SSC-Dampled-OMP}
\end{equation}
To efficiently solve problem in \eqref{eq:SSC-Dampled-OMP}, similar for OMP for problem \eqref{eq:SSC-OMP}, we also keep track of a working set $S^{(k)} \subseteq \{1, \cdots, N\}$, which is initialized as $S^{(0)} = \emptyset$, 
and incremented by one element at each iteration. Denote the current solution that is supported on $S^{(k)}$ as $\b_j^{(k)}$ and 
%
%
initialize the residual $\q_j^{(0)} = \x_j$. At the $k$-th iteration, we update $S^{(k)}$ and $\b_j^{(k)}$ to $S^{(k+1)}$ and $\b_j^{(k+1)}$, respectively,
by solving the following optimization problem
\begin{equation}
\label{eq:damped-index-greedy}
\min\limits_{b_{ij} \in \RR} \min\limits_{i\in\{1, \cdots, N\}} \|\q_j^{(k)} - \x_i b_{ij}\|_2^2 + \lambda (b_{ij}-c_{ij})^2.
\end{equation}
Note that
\begin{equation}
\label{eq:damped-select-index-i-star-objective}
\begin{split}
     &\|\q_j^{(k)} - \x_i b_{ij}\|_2^2 + \lambda (b_{ij}-c_{ij})^2 \\
    =& \|\q_j^{(k)}\|^2_2 - 2 \q_j^{(k)\top} \x_i b_{ij} +  (\|\x_i\|^2_2 + \lambda) b_{ij}^2 + \lambda c^2_{ij} - 2 \lambda b_{ij}c_{ij} \\
    =& \|\q_j^{(k)}\|^2_2 + (1 + \lambda) (b_{ij} -   \frac{\q_j^{(k)\top} \x_i + \lambda c_{ij}}{{ 1 + \lambda}})^2 \\
     &- \frac{(\q_j^{(k)\top} \x_i + \lambda c_{ij})^2 }{1+\lambda} + \lambda c^2_{ij}, \\
\end{split}
\end{equation}
where the last equality comes from the fact that $\| \x_i \|_2^2 =1$.
The optimal $i$ for problem \eqref{eq:damped-index-greedy} have a closed form as follows:
\begin{equation}
\label{eq:damped-index-i-star-appendex}
\begin{split}
i^\ast &=\arg \max_i \frac{(\x_i^\top\q_j^{(k)}  + \lambda c_{ij})^2 }{1+\lambda} - \lambda c^2_{ij} \\
       &=\arg \max_i \frac{(\x_i^\top \q_j^{(k)} + \lambda c_{ij})^2 - \lambda(1+\lambda) c^2_{ij}}{1+\lambda}\\
       &=\arg \max_i (\x_i^\top \q_j^{(k)})^2 + 2 \lambda \x_i^\top \q_j^{(k)}c_{ij} - \lambda c^2_{ij}.
\end{split}
\end{equation}
Since that there is no guarantee to avoid repeated index, we add an extra constraint $i \in \II \setminus S^{(k)}$ where $\II$ is the valid index set $\II$. Thus, we select the index $i^\ast$ as follows:
\begin{equation}
\label{eq:damped-index-i-star-without-repeatation}
\begin{split}
i^\ast &=\arg \max_{i \in \II \setminus S^{(k)}} (\x_i^\top \q_j^{(k)})^2 + 2 \lambda \x_i^\top \q_j^{(k)}c_{ij} - \lambda c^2_{ij}.
\end{split}
\end{equation}
Once such $i^\ast$ is selected, we update $S^{(k+1)}$ to $S^{(k)} \cup \{i^\ast\}$ and then update $\b_j$ by solving
\begin{equation}
\begin{split}
&\b_j^{(k+1)} = \mathop {\arg \min }\limits_{\b_j} \|\x_j - X \b_j\|_2^2 + \lambda \|\b_j - \c_j\|_2^2 \\
&\st \quad \text{supp}(\b_j) \subseteq S^{(k+1)},
\end{split}
\end{equation}
which has a closed-form solution
\begin{equation}
\label{eq:S^3COMP-Penalty-update-b-j-solution}
\begin{split}
&\arg \min_{\b^{[k+1]}_j} {\| {\x_j - X_{[k+1]} \b^{[k+1]}_j} \|^2_2}+\lambda \|\b^{[k+1]}_j - \c^{[k+1]}_j\|_2^2 \\
&= (X_{[k+1]}^\top X_{[k+1]} + \lambda \I)^{-1}(X_{[k+1]}^\top \x_j + \lambda \c^{[k+1]}_j),
\end{split}
\end{equation}
where $X_{[k+1]}$, $\b^{[k+1]}_j$ and $\c^{[k+1]}_j$ denote the submatrix or subvector of $X$, $\b^{(k+1)}_j$ and $\c_j$ corresponding to the support $S^{(k+1)}$, respectively, and $\I$ is an identity matrix with proper dimension. Then we update the residual $\q_j^{(k)}$ to $\q_j^{(k+1)} = \x_j - X \b_j^{(k+1)}$.


\subsection{Damped OMP with Picking $i^\ast$ from $\II \cup \J$}

%
%
%
To solve problem \eqref{eq:S$^3$COMP-Penalty-b-j}, a principled 
way is to update $\b_j$ from both the support in the index set $\II$ of the preserved columns 
and the support in the index set $\J$ of the dropped columns.

Again, we initialize the support set $S^{(0)}$ as an empty set, set the residual $\q_j^{(0)} = \x_j$, and find the support set $S^{(k+1)}$ of the solution $\b_j$ in a greedy search procedure 
by incrementing $S^{(k)}$ one index $i^\ast$ at each iteration.  


\begin{itemize}
\item {\bf Case I: $i^\ast \in \II$}

As derived in the previous section, we can compute $i^\ast$ as follows:
\begin{align}
\label{eq:damped-index-i-star-from-I}
i^\ast =\arg \max_{i \in \II \setminus S^{(k)} } \psi_i(\q_j^{(k)}, \c_j),
\end{align}
where $\psi_i(\q_j^{(k)}, \c_j) =(\x_i^\top \q_j^{(k)})^2 + 2 \lambda \x_i^\top \q_j^{(k)}c_{ij} - \lambda c^2_{ij}$. 
Note that the objective value in \eqref{eq:damped-index-greedy} of this case will be
\begin{align}
\|\q_j^{(k)}\|^2_2 - \frac{(\q_j^{(k)\top} \x_i + \lambda c_{ij})^2 }{1+\lambda} + \lambda c^2_{ij}.
\end{align}

\item {\bf Case II: $i^\ast \in \J$}

In this case, we consider the following problem:
\begin{equation}
\label{eq:damped-index-greedy-from-J}
\begin{split}
\min\limits_{b_{ij} \in \RR} \min\limits_{i \in \J} \|\q_j^{(k)} - \0 b_{ij}\|_2^2 + \lambda (b_{ij}-c_{ij})^2.
\end{split}
\end{equation}
It implies, from the first term in the objective function, that we can pick any $i^\ast \in \J$ and then set $b_{ij}=c_{ij}$. By doing so, the objective value in \eqref{eq:damped-index-greedy} of this case will be $\|\q_j^{(k)}\|_2^2$.

\end{itemize}

By comparing the objective values of picking $i^\ast$ via \eqref{eq:damped-index-i-star-from-I} and \eqref{eq:damped-index-greedy-from-J}, we conclude that if:
\begin{equation}
\begin{split}
\lambda c^2_{i^\ast j}- \frac{(\q_j^{(k)\top} \x_{i^\ast} + \lambda c_{i^\ast j})^2 }{1+\lambda} < 0
\end{split}
\end{equation}
holds, then 
$i^\ast$ will be picked via \eqref{eq:damped-index-i-star-from-I}; otherwise 
$i^\ast \in \J$ will be picked from \eqref{eq:damped-index-greedy-from-J}.
%
%
%
In the later case, we may select $i^\ast \in \J$ via \footnote{If we choose the index $i^\ast \in \J$ for which we have $c_{i^\ast j}=0$, then it leads to $b_{i^\ast j}=c_{i^\ast j}=0$. This choice will result in losing $1 \over s$ opportunity to pursuit a nonzero entry for $\b_j$ and thus result in $\| \b_j\|_0 \le s-1$.}
\begin{align}
\label{eq:damped-index-i-star-from-J}
i^\ast =\arg \max_{i \in \J \setminus S^{(k)} } | c_{ij} |.
\end{align}
%
%
%
%
%

By considering both $i^\ast \in \II$ and $i^\ast \in \J$ for picking $i^\ast$, the support of the optimal solution of the consensus problem over $T$ subproblems will be up to $s$. Nevertheless, unlike in SSCOMP \cite{You:CVPR16-SSCOMP}, the sparsity parameter $s$ can be larger than the dimension of the subspace due to the implicit squared $\ell_2$ norm regularization induced by the random dropout.

\myparagraph{Remark} Since that picking $i^\ast$ is a heuristic step in solving problem \eqref{eq:S$^3$COMP-Penalty-b-j}, rather than picking up $i^\ast$ from the index set $\J$ of the dropped columns, we only consider to pick up $i^\ast$ from the index set $\II$ of the preserved columns.

\section{More Experiments and Evaluations}
\label{sec:append-experiments-more}

In this section, we provide 
details for the evaluation metrics and present additional evaluation results.

\subsection{Evaluation Metrics}

To evaluate the performance and to illustrate the design idea, following the metrics used in \cite{You:CVPR16-SSCOMP}, we use clustering accuracy, subspace-preserving representation error, connectivity, and running time. The detailed definitions of the metrics are listed below.

\mysubparagraph{Clustering accuracy (acc: $a\%$)}
It is computed by matching the estimated clustering labels and the true labels as
\begin{align}
a  = \max\limits_{\pi} {100\over N}\sum_{i,j} Q_{\pi(i)j}^{est} Q_{ij}^{true},
\end{align}
where $\pi$ is a permutation of the $n$ groups, $Q^{est}$ and $Q^{true}$ are the estimated and ground-truth labeling of the data, respectively, with their $(i,j)$-th entry being equal to 1 if point $j$ belongs to cluster $i$ and zero otherwise.

\mysubparagraph{Subspace-preserving representation error (sre: $e\%$) \cite{Elhamifar:TPAMI13}}
For each $\c_j$, we compute the fraction of its $\ell_1$ norm that comes from other subspaces and then average over all $j$, \eg,
\begin{align}
e = \frac{100}{N}\sum_{j} (1 - \sum_i(\omega_{ij} \cdot |\c_{ij}|) / \|\c_j\|_1),
\end{align}
where $\omega_{ij} \in \{0,1\}$ is the true affinity. If $C=[\c_1,\cdots,\c_N]$ is subspace-preserving, we have $e = 0$.

\mysubparagraph{Connectivity (conn: $c$)} To evaluate the connectivity, we compute the \emph{algebraic connectivity}, which is defined as the second smallest eigenvalue of the normalized graph Laplacian.
For an undirected graph with weights $W \in \RR{ ^{N \times N}}$ and degree matrix $D = \text{diag} (W \cdot \mathbf{1})$, where
$\mathbf{1}$ is the vector of all ones, we use the second smallest eigenvalue $\lambda_2$ of the normalized Laplacian $L = I - D^{-1/2} W
D^{-1/2}$ to measure the connectivity of the graph, where $\lambda_2$ is in the range $[0, \frac{n-1}{n}]$, where $n$ is the number of clusters, and is zero if and only if the graph is not connected. The property of $\lambda_2$ is well summarized in \cite{Mohar:GTCA91}.
To evaluate the connectivity of the affinity graph with $n$ clusters, we compute the second smallest eigenvalue for each of the $n$ subgraphs and denote it as $\lambda_2^{(i)}$ for the $i$-th cluster. Then, 
the connectivity of the whole affinity graph is measured by
\begin{align}
c := \min_i \{\lambda_2^{(i)} \}_{i=1}^n.
\label{eq:algebraic-connectivity-c}
\end{align}
If $c=0$, then there is at least one affinity subgraph that is not connected.
We also define
\begin{align}
\bar c := \frac{1}{n}\sum_{i=1}^n \lambda_2^{(i)}
\label{eq:algebraic-connectivity-c-bar}
\end{align}
%
to measure the averaged algebraic connectivity.
For synthetic data, we use the quantity defined in \eqref{eq:algebraic-connectivity-c} because it is enough to show the difference in the connectivity; whereas for real world datasets, we also use the average quantity $\bar c$ defined in \eqref{eq:algebraic-connectivity-c-bar} in order to show the connectivity improvements on average especially when the quantity $c$ fails (\eg, $c=0$).
The most comprehensive and clear way to compare the connectivity of the $n$ subgraphs is to directly compare the corresponding $n$ second smallest eigenvalues $\{\lambda_2^{(i)}\}_{i=1}^n$ for the $n$ clusters computed from different methods, as in Fig. \ref{fig:GTSRB_conn_update}.

\mysubparagraph{Running time ($t$ (sec.))} For each subspace clustering task, we use 
MATLAB 2016. The reported numbers in all the experiments of this section are averages over 10 trials.

\subsection{More Experiments on Synthetic Data}
\label{sec:Synthetic Experiments_add}

\begin{figure}[thb]
\centering
\subfigure[{S}$^{3}$COMP]
{
    \includegraphics[width=0.420\columnwidth]{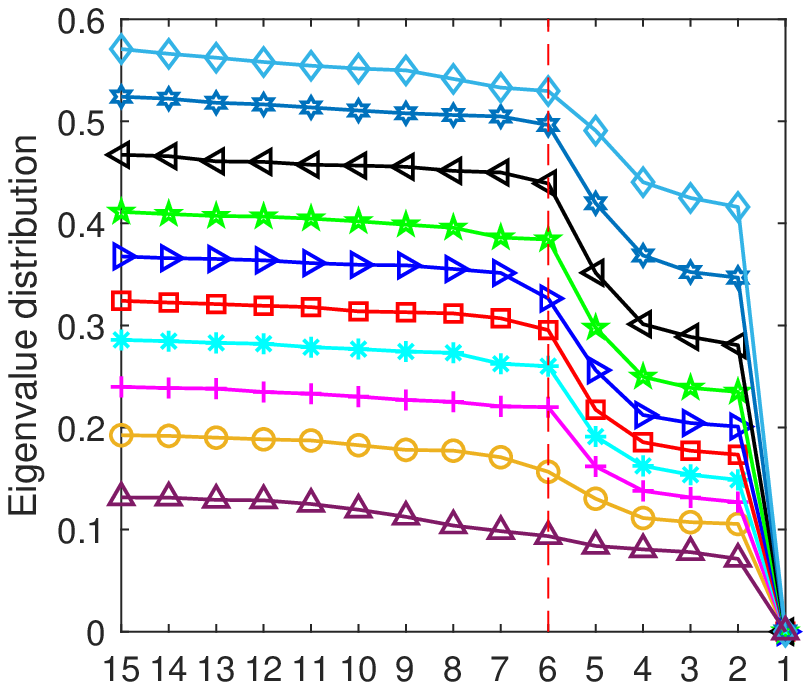}
    \label{figure:sub2-eig}
}
\hspace{-5.5mm}
\subfigure[{S}$^{3}$COMP-C]
{
    \includegraphics[width=0.575\columnwidth]{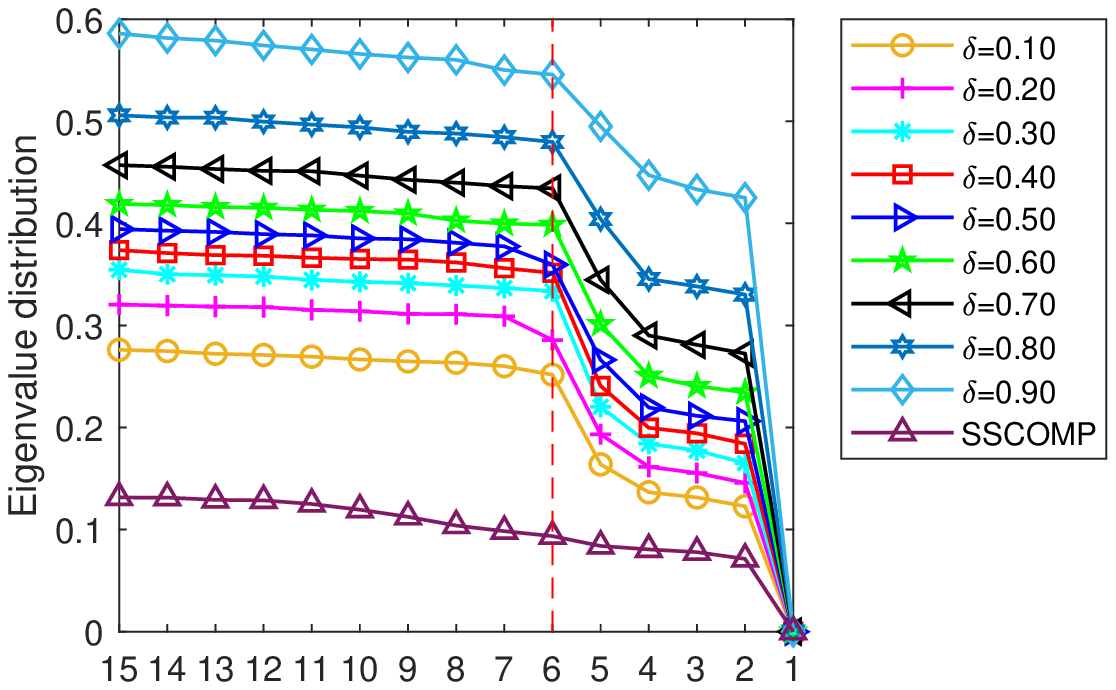}
    \label{figure:sub3-eig}
}
\caption{Minor eigenvalues of the normalized graph Laplacians under different $\delta$ on synthetic data with $N_i=320$.}
\label{fig:eigvalue-gap-vs-parameter-delta}
\end{figure}

\begin{figure}[bht]	
\centering
			\includegraphics[width=0.65\columnwidth]{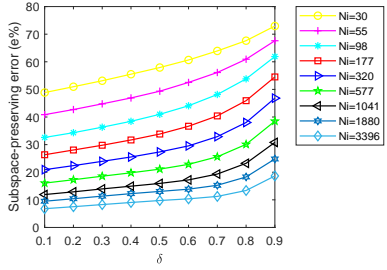}
			\label{fig:syn_consensus_delta_sse}
\caption{Subspace-preserving error of {S}$^{3}$COMP-C as functions of $\delta$ on synthetic data. } 
\label{fig:syn-result-acc-conn-sre-vs-delta-Ni-2}		
\end{figure}


\myparagraph{Effect of Improving the Connectivity}
In Fig.~\ref{fig:eigvalue-gap-vs-parameter-delta}, we show the last 15 minor eigenvalues of the graph Laplacian of S$^{3}$COMP and S$^{3}$COMP-C, compared to that of SSCOMP. Note that the number of subspaces is 5. If the graph Laplacian consists of five connected components, then there will be an eigenvalue gap between the ending fifth and sixth eigenvalues \cite{vonLuxburg:StatComp2007}. The larger the gap between the fifth and sixth eigenvalues is, the better the graph Laplacian becomes for spectral clustering. We can observe that all the eigenvalue curves of S$^{3}$COMP and S$^{3}$COMP-C under different parameter $\delta$ show shaper eigenvalue gaps; whereas the eigenvalue curves of SSCOMP does not have a clear eigenvalue gap.
This confirms that introducing the dropout into the self-expression model improves the connectivity. The improved connectivity will make the clustering patterns more clear and will thus help spectral clustering yield better clustering result.

\myparagraph{Effect of Dropout Rate $\delta$}
To evaluate the effect of the dropout rate $\delta$, we record the performance of {S}$^{3}$COMP-C with different dropout rate on synthetic data sets with different number of data points per subspace. We set $T=15$.
Experimental results are presented in Fig.~\ref{fig:syn-result-acc-conn-sre-vs-delta-Ni-2}.
As the dropout rate increases, the subspace-preserving error becomes larger. As shown in Fig.~\ref{fig:syn-result-acc-con-sre-vs-delta-Ni}(b), that the clustering accuracy tends to being stable.
This can be accounted from the improvements in the connectivity and the good tradeoff between the connectivity and the subspace-preserving errors. To be more specific, the improvements on the connectivity can help to avoid erroneous over-segmentation and at meantime it also helps to yield more compact clusters in spectral embedding---both sides help to obtain better clustering result.

\subsection{More Experiments on Real World Data}
\label{sec:Real-world-data-experiments-add}

\myparagraph{Comparing Connectivity}
In Table \ref{tbl:comparison-conn-mean-min-real-datasets}, we compare the connectivity of the affinity graph induced by different methods on each real world dataset. Note that {S}$^{3}$COMP is a simplified version of {S}$^{3}$COMP-C, that performs only one iteration; whereas {S}$^{3}$COMP$^\ast$ differs from {S}$^{3}$COMP in the step of computing the consensus solution $\c_j$ from $\{\b_j^{(t)}\}_{t=1}^T$. Specifically, rather than taking the average by dividing $T$, we compute the nonzero entry $c_{ij}$ via $\frac{1}{r_i}\sum_{t=1}^T b_{ij}^{(t)}$, where $r_i$ is the number of the nonzero entries in $\{b_{ij}^{(t)}\}_{t=1}^T$. 

Table \ref{tbl:comparison-conn-mean-min-real-datasets} shows that, the affinity graphs produced by {S}$^{3}$COMP and {S}$^{3}$COMP-C have 
nonzero connectivity $c$ on MNIST4000, MNIST10000, MNIST70000, and GTSRB.
%
Although the connectivity $c$ for S$^3$COMP and S$^3$COMP-C is zero on Extended Yale B and COIL100, the averaged connectivity $\bar{c}$ has a significant improvement over that for SSCOMP.



\begin{table}[hbt]
\centering
\footnotesize
\begin{threeparttable}
\resizebox{0.5\textwidth}{!}{
\begin{tabular}{l|c c c c c c}
\hline
   conn                      & SSCOMP & EnSC  & {S}$^{3}$COMP$^\ast$ & {S}$^{3}$COMP & {S}$^{3}$COMP-C \\
\hline
\multicolumn{5}{c}{ExYaleB} \\
\hline
min:  $ c$              & 0.0000 & 0.0401& 0.0102               & 0.0000        & 0.0000  \\
mean:  $\bar c$          & 0.0381 & 0.0550& 0.0749               & 0.0726        & 0.0667  \\
\hline
\multicolumn{5}{c}{COIL100} \\
\hline
min:  $ c$              & 0.0000 & 0.0000& 0.0000               & 0.0000        & 0.0000  \\
mean:  $\bar c$          & 0.0060 & 0.0163& 0.0213               & 0.0081        & 0.0077  \\
\hline
\multicolumn{5}{c}{MNIST4000} \\
\hline
min:  $ c$              & 0.0491 & 0.0518& 0.1325               & 0.0954        & 0.0953  \\
mean:  $\bar c$          & 0.1371 & 0.1117& 0.1883               & 0.1529        & 0.1527  \\
\hline
\multicolumn{5}{c}{MNIST10000} \\
\hline
min:  $ c$              & 0.0205 & 0.0412& 0.1241               & 0.1174        & 0.1174  \\
mean:  $\bar c$          & 0.1212 & 0.0938& 0.1824               & 0.1720        & 0.1719  \\
\hline
\multicolumn{5}{c}{MNIST70000} \\
\hline
min:  $ c$              & 0.0000 & 0.0000& 0.0911               & 0.1026        & 0.1026  \\
mean:  $\bar c$          & 0.0830 & 0.0911 & 0.1646               & 0.1569        & 0.1569  \\
\hline
\multicolumn{5}{c}{GTSRB} \\
\hline
min:  $ c$              & 0.0000 & 0.0000 & 0.0114               & 0.0051        & 0.0051  \\
mean:  $\bar c$          & 0.0213 & 0.0095 & 0.1883               & 0.0576        & 0.0573  \\
\hline
\end{tabular}
}
\\[3pt]
\end{threeparttable}
\caption{Connectivity comparison for SSCOMP, {S}$^{3}$COMP and {S}$^{3}$COMP-C on real world dataset. The connectivity metric $c$ is defined in \eqref{eq:algebraic-connectivity-c} and the connectivity metric $\bar c$ is defined in \eqref{eq:algebraic-connectivity-c-bar}}
\label{tbl:comparison-conn-mean-min-real-datasets}
\end{table}

\begin{figure*}[bht]	
\centering
		\subfigure[]
		{
			\includegraphics[width=0.880\columnwidth]{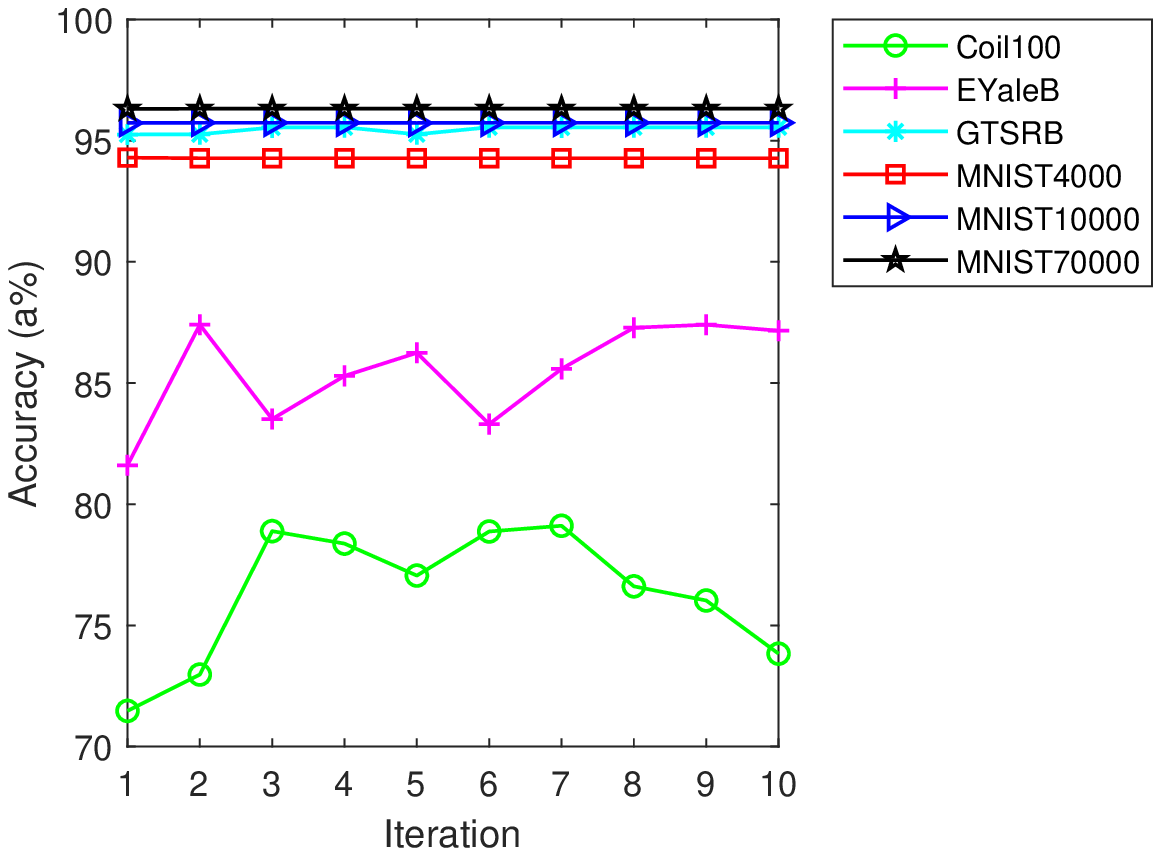}
			\label{fig:real_consensus_iter_acc}
		}
		\hspace{-5.5mm}
		\subfigure[]
		{
			\includegraphics[width=0.6415\columnwidth]{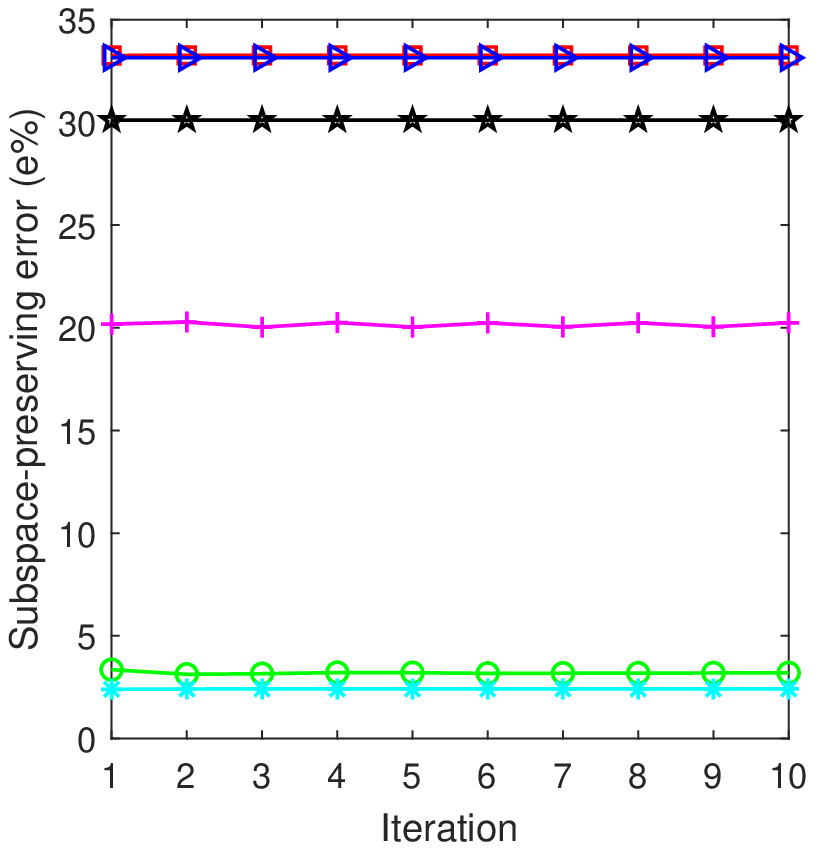}
			\label{fig:real_consensus_iter_sre}
		}
\caption{Clustering accuracy and subspace-preserving error of {S}$^{3}$COMP-C as functions of iterations on real world data.}
\label{fig:real-consensus-iter-acc-sre}		
\end{figure*}

\begin{figure*}[bht]	
\centering
\subfigure[EYaleB]
		{
			\includegraphics[width=0.65\columnwidth]{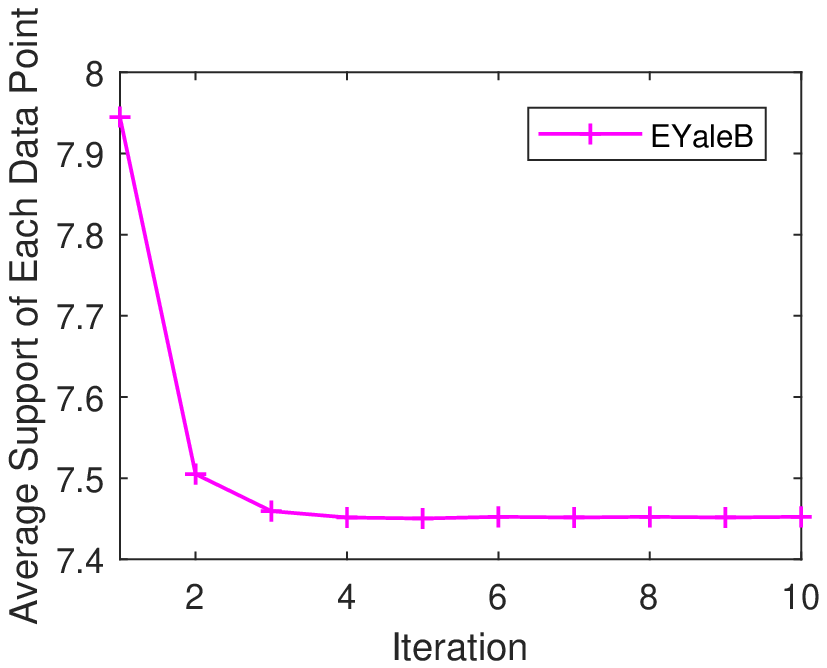}
			\label{fig:supp-vs-iteration-in-consensus-real-data-EYaleB}
		}
		\subfigure[COIL100]
		{
			\includegraphics[width=0.65\columnwidth]{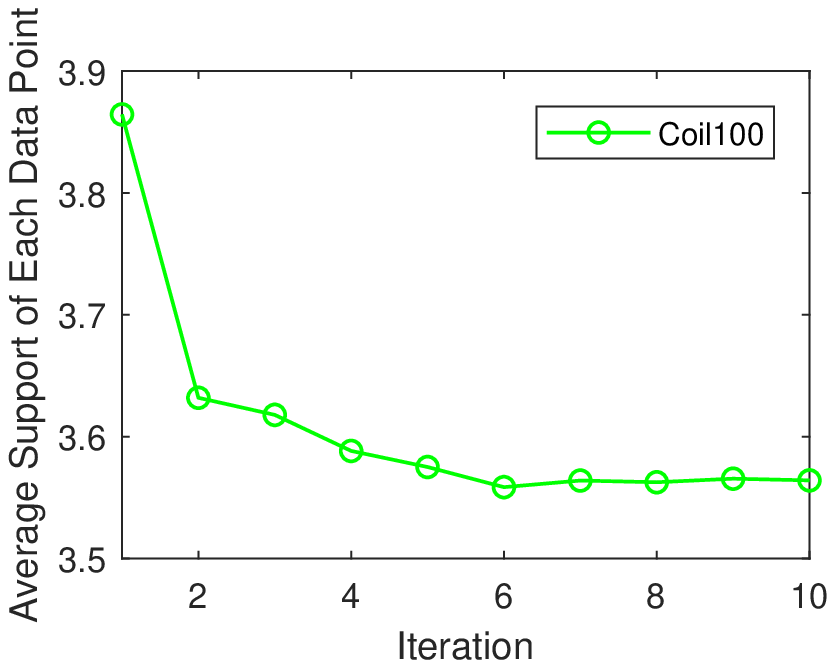}
			\label{fig:supp-vs-iteration-in-consensus-real-data-coil100}
		}
		\subfigure[GTSRB]
		{
			\includegraphics[width=0.65\columnwidth]{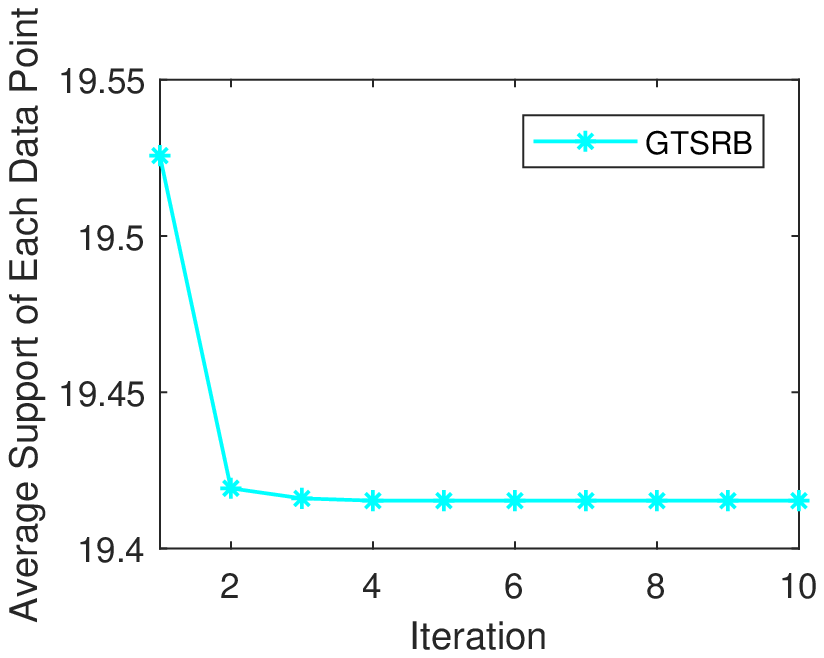}
			\label{fig:supp-vs-iteration-in-consensus-real-data-GTSRB}
		}
\\
		\subfigure[MNIST4000]
		{
			\includegraphics[width=0.65\columnwidth]{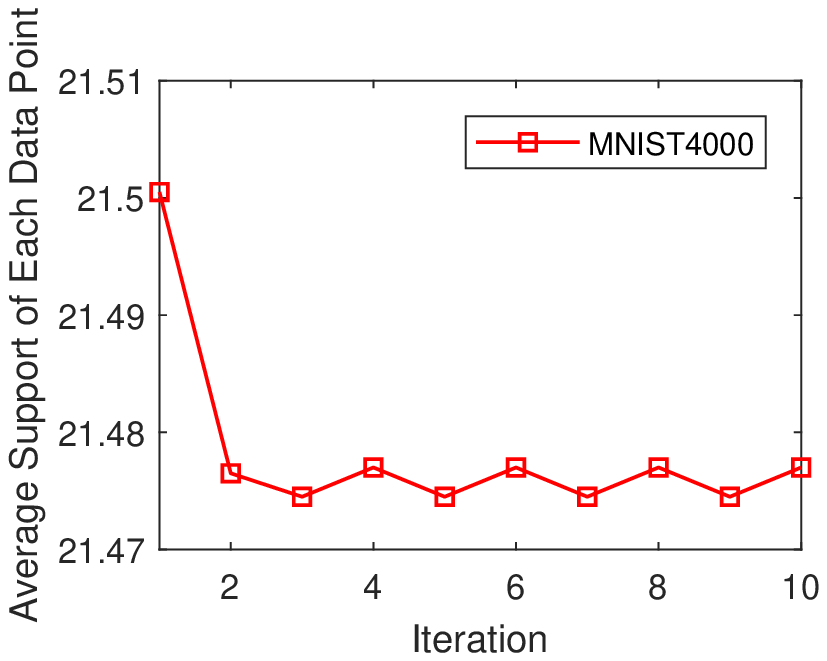}
			\label{fig:supp-vs-iteration-in-consensus-real-data-MNIST4000}
		}
		\subfigure[MNIST10000]
		{
			\includegraphics[width=0.65\columnwidth]{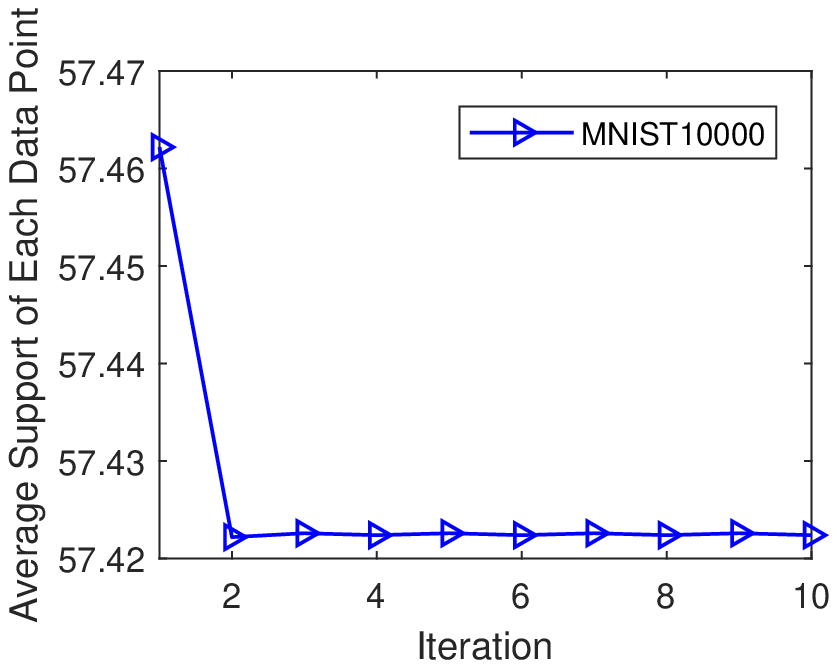}
			\label{fig:supp-vs-iteration-in-consensus-real-data-MNIST10000}
		}
		\subfigure[MNIST70000]
		{
			\includegraphics[width=0.65\columnwidth]{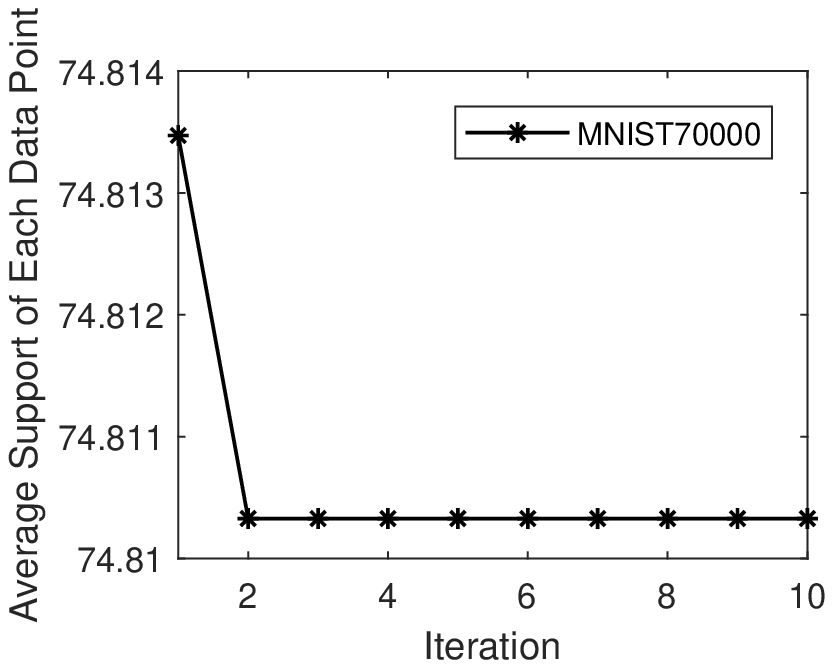}
			\label{fig:supp-vs-iteration-in-consensus-real-data-MNIST70000}
		}

\caption{Support changes during iterations in consensus of {S}$^{3}$COMP-C on real world data.}
\label{fig:support-changes-vs-iterations-on-real-data}		
\end{figure*}

\myparagraph{Performance Changes during the Outer Iterations of S$^3$COMP-C}
To evaluate the performance changes during the outer iteration of S$^3$COMP-C, we conduct experiments on the real world dataset where the maximum number of the outer iterations is set to $10$, and display the clustering accuracy, subspace-recovering representation errors, and the support size as a function of the iteration number, respectively.
%
%
In Fig. \ref{fig:real-consensus-iter-acc-sre}, we observe that the subspace-preserving error and the clustering accuracy are stable, except for the clustering accuracy on Extended Yale B and COIL100 showing some fluctuations.
%
%
We calculate the average size of the supports of the returned consensus solutions on each real world dataset, and present them as a function of the outer iteration number in Fig.~\ref{fig:support-changes-vs-iterations-on-real-data}. The average support size on each dataset 
becomes stable in very few number of iterations.

\subsection{Setting the Dropout Parameter $\delta$ and the Penalty Parameter $\lambda$}
\label{sec:parameters-in-experiments}
%

%
%
%
%
We have some guiding principles for setting the parameters. For the dropout rate $\delta$, we recommend picking a larger $\delta$ for larger datasets. For the sparsity $s$, it should be similar to the dimension of the subspaces, for which there could be a rough estimate in certain applications. The sparsity $s$ in SSCOMP relates to the intrinsic dimension of subspace. To be fair, we set $s$ to be the same as that in SSCOMP. 
For the penalty parameter $\lambda$, a proper range can be determined by checking the sharpness of the ``eigenvalue gap'', as illustrated in Fig.~\ref{fig:eigvalue-gap-vs-parameter-delta}. Experiments show that the results are less sensitive to $\lambda$ whenever it is in \emph{the proper range} in where the connectivity can be notably improved, \eg, [0.1,1.0] (except for EYaleB, which is (0, 0.05]). 
%
%
%
We listed the parameters $\lambda$ in problem \eqref{eq:S^3COMP-Penalty} and $\delta$ used on each dataset in Table~\ref{tab:parameters-used-in-synthetic-data-set} and Table~\ref{tab:parameters-used-in-real-world-data-set}.

\begin{table}[thb]
\centering
\small
\resizebox{0.475\textwidth}{!}{
\begin{tabular}{l|c c c c c c c c c}
\hline
$N_i  $                & 30   & 55   &  98   & 177  & 320  & 577  & 1041 & 1880 & 3396 \\
\hline
$\lambda$ & 0.40 & 0.40 &  0.70 & 0.70 & 0.70 & 1.00 & 1.00 & 1.00 & 1.00 \\
$\delta$  & 0.30 & 0.30 &  0.30 & 0.40 & 0.40 & 0.60 & 0.60 & 0.60 & 0.60 \\

\hline
\end{tabular}
}
\vspace{3pt}
\caption{Parameters $\delta$ and $\lambda$ used in S$^3$COMP-C on synthetic datasets}
\label{tab:parameters-used-in-synthetic-data-set} 
\end{table}

\begin{table*}[thb]
\centering
\small
\resizebox{0.750\textwidth}{!}{
\begin{tabular}{l|c c c c c c}
\hline
Datasets                & Ex. Yale B   & COIL100   &  MNIST4000  & MNIST10000  & MNIST70000 & GTSRB  \\
\hline
                         $\lambda$ & 0.01 & 0.60 &  0.10 & 1.00 & 0.80 & 0.80  \\
                         $\delta$    & 0.10 & 0.10 &  0.10 & 0.60 & 0.80 & 0.80  \\
\hline
\end{tabular}
}
\vspace{3pt}
\caption{Parameters $\delta$ and $\lambda$ used in S$^3$COMP-C on real world datasets}
\label{tab:parameters-used-in-real-world-data-set} 
\end{table*}

{\small
\bibliographystyle{ieee_fullname}
\bibliography{biblio//cgli,biblio//deeplearning,biblio//sparse,biblio//learning,biblio//vidal,biblio//geometry,biblio//segmentation}
}

\end{document}